\documentclass{article}
\usepackage[a4paper, total={6.5in, 9in}]{geometry}

\usepackage{amsmath,amssymb,pifont}
\usepackage{multicol}
\usepackage{amstext}
\usepackage{amsthm}
\usepackage{multirow}
\usepackage{booktabs}
\usepackage[skip=0pt]{subcaption}
\usepackage{authblk}
\usepackage{lipsum}
\usepackage[shortlabels]{enumitem}
\usepackage{cancel}
\usepackage{array}
\usepackage{siunitx}
\usepackage{csvsimple}
\usepackage[multidot]{grffile}
\usepackage{bbm}
\usepackage{hyperref}
\usepackage{makecell}
\usepackage{bbm, dsfont}
\usepackage{mathtools}
\usepackage{xcolor}
\usepackage{comment}
\usepackage{natbib}
\usepackage{footnote}

\usepackage{mathrsfs}
\usepackage{microtype}
\usepackage{graphicx}
\usepackage[capitalize,noabbrev]{cleveref}
\usepackage{url}

\usepackage{algorithm}
\usepackage{algorithmic}
\usepackage{xfrac}

\makeatletter
\newcommand{\specificthanks}[1]{\@fnsymbol{#1}}%
\makeatother

\newtheorem{lemma}{Lemma}

\hypersetup{final}

\newcommand{\eps}{\ensuremath{\varepsilon}}

\newcommand{\calA}{\ensuremath{\mathcal{A}}}

\newcommand{\calC}{\ensuremath{\mathcal{C}}}

\newcommand{\calL}{\ensuremath{\mathcal{L}}}

\newcommand{\calN}{\ensuremath{\mathcal{N}}}
\newcommand{\calO}{\ensuremath{\mathcal{O}}}
\newcommand{\calP}{\ensuremath{\mathcal{P}}}
\newcommand{\calQ}{\ensuremath{\mathcal{Q}}}
\newcommand{\calR}{\ensuremath{\mathcal{R}}}

\renewcommand{\Pr}{\mathop{\mathbf{Pr}}}

\newtheorem{lem}{Lemma}[section]

\newtheorem{thm}[lem]{Theorem}

\newtheorem{defn}[lem]{Definition}

\newtheorem{definition}[lem]{Definition}

\DeclareMathOperator*{\argmax}{arg\,max}

\makeatletter
\newcommand{\vast}{\bBigg@{4}}
\newcommand{\Vast}{\bBigg@{5}}
\makeatother

\newcommand{\ex}[2]{{\ifx&#1& \mathbb{E} \else
\underset{#1}{\mathbb{E}} \fi \left[#2\right]}}
\newcommand{\pr}[2]{{\ifx&#1& \mathbb{P} \else
\underset{#1}{\mathbb{P}} \fi \left[#2\right]}}

\newcommand{\Var}[1]{\ensuremath{\mathbf{Var}\left(#1\right)}}
\newcommand{\Cov}[2]{\ensuremath{\mathbf{Cov}\left(#1, #2\right)}}

\newcommand{\ltwo}[1]{\left\|#1\right\|_2}
\newcommand{\ltwosq}[1]{\left\|#1\right\|_2^2}

\newcommand{\algtr}[1]{${\sf #1}_{\sf tr}$}
\newcommand{\alginf}[1]{${\sf #1}_{\sf inf}$}

\newcommand{\mypar}[1]{\smallskip
	\noindent{\textbf{{#1}:}}}
	
\renewcommand{\epsilon}{\varepsilon}

\renewcommand{\tilde}{\widetilde}

\newcommand{\privTa}{\theta^{\sf UTA}_{\sf priv}\,}

\newcommand{\hmu}{\widehat{\mu}}
\newcommand{\emaBase}{{\sf EMA}_{\sf baseline}}

\newcommand{\expected}[2]{\underset{#2}{\mathbb{E}}[#1]}

\begin{document}

\title{Recycling Scraps: Improving Private Learning \\ by Leveraging Checkpoints}

\author{
  Virat Shejwalkar\thanks{Work done while the author was an intern at Google.}, Arun Ganesh\thanks{Listed in alphabetical order.}, Rajiv Mathews\textsuperscript{\specificthanks{2}}, Yarong Mu\textsuperscript{\specificthanks{2}}, Shuang Song\textsuperscript{\specificthanks{2}}, Om Thakkar\textsuperscript{\specificthanks{2}}, Abhradeep Thakurta\textsuperscript{\specificthanks{2}}, Xinyi Zheng\textsuperscript{\specificthanks{2}} \\
  Google\\
  \{vshejwalkar, arunganesh, mathews, ymu, shuangsong, omthkkr, athakurta, cazheng\}@google.com
}

\date{}

\maketitle

\begin{abstract}
In this work, we focus on improving the  accuracy-variance  trade-off for state-of-the-art differentially private machine learning (DP ML) methods. First, we design a general framework that uses aggregates of intermediate checkpoints \emph{during training} to increase the accuracy of DP ML techniques. Specifically, we demonstrate that training over aggregates can  provide significant gains in prediction accuracy over the existing state-of-the-art for StackOverflow, CIFAR10 and CIFAR100 datasets. For instance, we improve the state-of-the-art DP StackOverflow accuracies to 22.74\% (+2.06\% relative) for $\epsilon=8.2$, and 23.90\%  (+2.09\%) for $\epsilon=18.9$.  Furthermore, these gains magnify in settings with periodically varying training data distributions. 
We also demonstrate that our methods achieve relative improvements of 0.54\% and 62.6\% in terms of utility and variance, on a proprietary, production-grade pCVR task.
Lastly, we initiate an exploration into estimating the uncertainty (variance) that DP noise adds in the predictions of DP ML models. We prove that, under standard assumptions on the loss function, the sample variance from last few checkpoints provides a good approximation of the variance of the final model of a DP run. Empirically, we show that the last few checkpoints can provide a reasonable lower bound for the variance of a converged DP model. Crucially, all the methods proposed in this paper operate on \emph{a single training run} of the DP ML technique, thus incurring no additional privacy cost.
\end{abstract}

\section{Introduction}\label{new_intro}

Machine learning models can unintentionally memorize sensitive information about the data they were trained on, which has led to numerous attacks that extract private information about the training data~\citep{ateniese2013hacking,fredrikson2014privacy,fredrikson2015model,carlini2019secret,shejwalkar2021membership,carlini2021extracting,carlini2022membership}. 
For instance, membership inference attacks~\citep{shokri2017membership} can infer whether a target sample was used to train a given ML model, while property inference attacks~\citep{melis2019exploiting,mahloujifar2022property} can infer certain sensitive properties of the training data. 
To address such privacy risks, literature has introduced various approaches to privacy-preserving ML~\citep{nasr2018machine,shejwalkar2021privacy,tang2022mitigating}. In particular, iterative techniques like differentially private stochastic gradient descent (DP-SGD)~\citep{song2013stochastic, BST14, DP-DL,mcmahan2017learning} and DP Follow The Regularized Leader (DP-FTRL)~\citep{kairouz2021practical} have become the state-of-the-art for training DP neural networks. 

The accuracy-variance trade-off is a central problem in machine learning.
Note that here, we use the term accuracy to refer to the primary evaluation metric of a model on the the training/test data sets, e.g., accuracy for datasets like CIFAR10 and StackOverflow, and AUC-loss (i.e., 1 - AUC) for datasets like pCVR.
Techniques like DP-SGD and DP-FTRL involve the operation of \emph{per-example gradient clipping} and calibrated Gaussian noise addition in each training step, which makes this trade-off even trickier to understand in DP ML~\cite{song21clipping}.
In this work, we focus on both fronts of the problem.

\mypar{Our contributions at a glance} 
First, we design a general  framework that (adaptively) uses aggregates of intermediate checkpoints (i.e., the intermediate iterates of model training) to increase the accuracy of DP ML techniques.
Next, we provide a method to estimate the uncertainty (variance) that DP noise adds to DP ML training.
Crucially, we attain both these goals with \emph{a single training run} of the DP technique, thus incurring no additional privacy cost.
While both the goals are interleaved, for ease of presentation, we will separate the exposition into two parts. In the following, we provide the details of our contributions, and place them in the context of prior works.

\mypar{Increasing accuracy using checkpoint aggregates (Sections~\ref{acc} and~\ref{sec:empEval})}
While the privacy analyses for state-of-the-art DP ML techniques allow releasing/using all the training checkpoints, prior works in DP ML~\citep{DP-DL,mcmahan2017learning,mcmahan2018general,PAS,WangBK19,zhu2019poission,balle2020privacy,esa++,papernot2020tempered,tramer2020differentially,andrew2021differentially,kairouz2021practical,amid2022public,feldman2022hiding}  use only the final model output by the DP algorithm for establishing benchmarks.
This is also how DP models are deployed in practice~\citep{ramaswamy2020training,google_dp}.
To our knowledge, \cite{de2022unlocking} is the only prior work that re-uses intermediate checkpoints to increase the accuracy of DP-SGD. They note non-trivial accuracy gains  by post-processing the  DP-SGD checkpoints using an exponential moving average (EMA). 
While \citep{chen2017checkpoint,izmailov2018averaging} explore checkpoint aggregation methods to improve performance in (non-DP) ML settings, they observe negligible performance gains.

In this work, \emph{we propose a general framework that adaptively uses intermediate checkpoints to increase the accuracy of state-of-the-art DP ML techniques}.
To our knowledge, this is the first work to re-use intermediate checkpoints \emph{during DP ML training}.
Empirically, we demonstrate significant performance gains using our framework for a next word prediction task with user-level DP for StackOverflow, an image classification task with sample-level DP for CIFAR10, and an ad-click conversion prediction task with sample-level DP for a proprietary pCVR dataset.
It is worth noting that DP state-of-the-art for benchmark datasets has repeatedly improved over the years since the foundational techniques from \cite{DP-DL} for CIFAR10 and \cite{mcmahan2017learning} for StackOverflow, hence any consistent improvements are instrumental in advancing the state of DP ML. 

Specifically, we show that training over aggregates of checkpoints achieves state-of-the-art prediction accuracy of 22.74\% at $\epsilon=8.2$ for StackOverflow (i.e., 2.09\% relative gain over DP-FTRL from \cite{kairouz2021practical})\footnote{These improvements are notable since there are 10$k$ classes in StackOverflow data.}, and 57.51\% at $\epsilon=1$ for CIFAR10 (i.e., 2.7\% relative gain over DP-SGD as per \cite{de2022unlocking}), respectively. {For CIFAR100 task, we first improve the DP-SGD baseline of~\cite{de2022unlocking} even without using any of our aggregation methods. Similar to \cite{de2022unlocking}, we warm-start DP training on CIFAR100 from a checkpoint pre-trained on ImageNet. However, we use the EMA checkpoint of the pre-training pipeline instead of the last checkpoint as in~\cite{de2022unlocking}, and improve DP-SGD performance by 5\% and 3.2\% for $\epsilon$ 1 and 8, respectively. Next, we show that training over aggregates further improves the accuracy on CIFAR100 by 0.67\% to 76.18\% at $\epsilon=1$ (i.e., 0.89\% relative gain over our improved CIFAR100 DP-SGD baseline).
}
Next, we show that these benefits further magnify in more practical settings with periodically varying training data distributions.
For instance, we note relative accuracy gains of 2.64\% and 2.82\% for $\epsilon$ of 18.9 and 8.2, respectively, for StackOverflow over DP-FTRL baseline in such a setting.
We also experiment with a proprietary, production-grade pCVR dataset~\cite{denison2022private,chua2024training} and show that at $\epsilon=6$, training over aggregates of checkpoints improves AUC-loss (i.e., 1 - AUC) by 0.54\% (relative) over the DP-SGD baseline.  Note that such an improvement is considered very significant in the context of ads ranking.
Theoretically, we show in Theorem~\ref{thm:tail} that for standard training regimes, the excess empirical risk of the final checkpoint of DP-SGD is $\text{log}(n)$ times more than that of the weighted average of the past $k$ checkpoints, where $n$ is the size of dataset.
It is interesting to theoretically analyze the use of checkpoint aggregations during training, which we leave as future work.  

\mypar{Uncertainty quantification using intermediate checkpoints (Section~\ref{sec:uncertainty})}
There are various sources of randomness in an ML training pipeline~\citep{abdar2021review}, e.g., choice of initial parameters, dataset, batching, etc. 
This randomness induces uncertainty in the predictions made using such ML models. 
In critical domains, e.g., medical diagnosis, self-driving cars and financial market analysis, failing to capture the uncertainty in such predictions can have  undesirable repercussions. DP learning adds an additional source of randomness by injecting noise at every training round. 
Hence, it is paramount to quantify reliability of the DP models, e.g., by quantifying the uncertainty in their predictions.

As prior work, \cite{karwa2017finite} develop finite sample confidence intervals but for the simpler Gaussian mean estimation problem.
Various methods exist
for uncertainty quantification in ML-based systems~\citep{mitchell1980need,roy2018inherent,begoli2019need,hubschneider2019calibrating,mcdermott2019deep,tagasovska2019single,wang2019aleatoric,nair2020exploring,ferrando2022parametric}.
However, these methods either use specialized (or simpler) model architectures to facilitate uncertainty quantification, or are not directly applicable to quantify the uncertainty in DP ML due to DP noise. 
For example, a common way of uncertainty quantification~\citep{barrientos2019differentially,nissim2007smooth,brawner2018bootstrap,evans2020statistically} that we call the \emph{independent runs} method, needs $k$ independent (bootstrap) runs of the ML algorithm. 
However, repeating a DP ML algorithm multiple times can incur significant privacy and computation costs.

To this end,~\emph{for the first time we quantify the uncertainty that DP noise adds} to DP training procedure~\emph{using only a single training run}.
We propose to use the last $k$ checkpoints of a single run of a DP ML algorithm as a proxy for the $k$ final checkpoints from independent runs. 
This does not incur \emph{any additional privacy cost} to the DP ML algorithm. 
Furthermore, it is useful in practice as it does not incur additional training compute, and can work with any algorithm having intermediate checkpoints. Finally, it doesn't require changing the underlying model or algorithm, unlike some other methods for uncertainty estimation (e.g., the use of Bayesian neural networks \cite{Zhang21Bayesian}).

Theoretically, we consider using (a rescaling of) the sample variance of a statistic $f(\theta)$ at checkpoints $\theta_{t_1}, \ldots, \theta_{t_k}$ as an estimator of the variance of any convex combination of $f(\theta_{t_i})$, i.e., any weighted average of the statistics at the checkpoints, and give a bound on the bias of this estimator. 
As expected, our bound on the error decreases as the ``\emph{burn-in}'' time $t_1$ and the time between checkpoints $t_2$ both increase. 
An upshot of this analysis is that getting $k$ nearly i.i.d. checkpoints requires fewer iterations than running $k$ independent runs of $t_1$ iterations. ~\emph{In turn, under a fixed privacy constraint, using the sample variance of the checkpoints can provide more samples and thus tighter confidence intervals than the independent runs method}; see the remark in Section~\ref{sec:uncertainty} for details.

Intuitively, our proof shows that (i) as the burn-in time increases, the marginal distribution of each $\theta_{t_i}$ approaches the distribution of $\theta_{t_k}$, and (ii) as the time between checkpoints increases, any pair $\theta_{t_i}, \theta_{t_j}$ approaches pairwise independence. We prove both (i) and (ii) via a mixing time bound, which shows that starting from any point distribution $\theta_0$, the Markov chain given by DP-SGD approaches its stationary distribution at a certain rate. 

Empirically, we show that our method provides reasonable lower bounds on the uncertainty quantified using the more accurate (but privacy and computation intensive) method that uses independent runs. 
For instance, we show that for DP-FTRL trained StackOverflow, the 95\% confidence widths for the scores of the predicted labels computed using independent runs method (no budget split)\footnote{Thus, a superior baseline by not splitting the privacy budget among the independent runs.} are always within a factor of 2 of the widths provided by our method for various privacy levels and number of bootstrap samples.

While we compute the variance in regards to a fixed prediction function, we believe our estimator can be used to obtain DP parameter confidence intervals for traditional statistical estimators (\emph{e.g.,} linear regression). We leave this direction for future exploration.

\section{Background and Preliminaries}\label{background}

In this section, we briefly introduce the background on machine learning, privacy leakages in machine learning models, differential privacy and deep learning with differential privacy.

\subsection{Machine Learning}\label{background:ml}
In this paper, we consider machine learning (ML) models used for image classification and language next-word-prediction tasks. We use \emph{supervised machine learning} for both the types of tasks and briefly review it below.

Let $f_{\theta}:\mathbb{R}^d \mapsto \mathbb{R}^k$ be a ML classifier (e.g., neural network) with $d$ input features and $k$ classes, which is parameterized by $\theta$. For a given example $\textbf{z} = (\textbf{x}, y)$, $f_{\theta}(\textbf{x})$ is the classifier's confidence vector for $k$ classes and the predicted label is the corresponding class which has the largest confidence score, i.e., $\hat{y} =  \mathop{\argmax}_i f_{\theta}(\textbf{x})$.
The goal of supervised machine learning is to learn the relationship between features and labels in given \emph{labeled} training data $D^l_{tr}$ and generalize this ability to unseen data. The model learns this relationship using empirical risk minimization (ERM) on the training set $D^l_{tr}$, where the risk is measured in terms of a certain loss function, e.g., cross-entropy loss:
$$\min_{\theta}\frac{1}{|D^l_{tr}|}\sum_{\textbf{z}\in D^l_{tr}}l(f_{\theta}, \textbf{z})\big)$$
Here $|D^l_{tr}|$ is the size of the labeled training set and $l(f_{\theta}, \textbf{z})$ is the loss function.
When clear from the context, we use $f$ instead of $f_\theta$, to denote the target model.

\subsection{Privacy Leakage in ML Models}\label{background:privacy_leakage_ml}
ML models generally require large amounts of training data to achieve good performances. This data can be of sensitive nature, e.g., medical records and personal photographs, and without proper precautions, ML models may leak sensitive information about their private training data. Multiple previous works have demonstrated this via various \emph{inference} attacks, e.g., membership inference, property or attribute inference, model stealing, and model inversion. Below, we review these attacks.

Consider a target model $f_{\theta}$ trained on $D_{tr}$ and a target sample $(\mathbf{x},y)$. Membership inference attacks~\cite{shokri2017membership,sankararaman2009genomic,ateniese2015hacking} aim to infer whether the target sample $(\mathbf{x},y)$ was used to train the target model, i.e., whether $(\mathbf{x},y)\in D_{tr}$. Property or attribute inference attacks~\cite{melis2019exploiting,song2019overlearning} aim to infer certain attributes of $(\mathbf{x},y)$ based on model's inference time representation of $(\mathbf{x},y)$. For instance, even if $f_{\theta}$ is just a gender classifier, $f_{\theta}(\mathbf{x})$ may reveal the race of the person in $\mathbf{x}$. Model stealing attacks~\cite{tramer2016stealing,orekondy2019knockoff} aim to reconstruct the parameters $\theta$ of the original model $f_{\theta}$ based on black-box access to $f_{\theta}$, i.e., using $f_{\theta}(\mathbf{x})$. Model inversion attacks~\cite{fredrikson2015model} aim to reconstruct the whole training data $D_{tr}$ based on white-box, i.e., using $\theta$, or black-box, i.e., using $f_{\theta}(\mathbf{x})$, access to model.

\subsection{Deep Learning with Differential Privacy}\label{background:dpsgd}
Differential privacy~\cite{DMNS,dwork2008differential,dwork2014algorithmic} is a notion to quantify the privacy leakage from the outputs of a data analysis procedure and is the gold standard for data privacy.
It is formally defined as below:

\begin{definition}[{{Differential Privacy}}]
    A randomized algorithm  $\mathcal{M}$ with domain $\mathcal{D}$ and range $\mathcal{R}$ preserves $(\varepsilon,\delta)$-differential privacy iff for any two neighboring   datasets $D,D' \in \mathcal{D}$ and for any subset $S \subseteq \mathcal{R}$ we have:
    
    \begin{align}
        \Pr[\mathcal{M}(D) \in S] \leq e^{\varepsilon} \Pr[\mathcal{M}(D') \in S] + \delta
    \end{align}\label{def:dp}
    
   where $\varepsilon$ is the \emph{privacy budget} and $\delta$ is the \emph{failure probability}. 
\end{definition}

R\'enyi Differential Privacy~(RDP) is a commonly-used relaxed definition for differential privacy. 

\begin{definition}[{{R\'enyi Differential Privacy~(RDP)}}~\cite{mironov2017renyi}]
 A randomized algorithm  $\mathcal{M}$ with domain $\mathcal{D}$ is $(\alpha,\varepsilon)$-RDP with order $\alpha \in (1,\infty)$
if and only if for any two   neighboring datasets $D,D'\in \mathcal{D}$:
\begin{align}
\label{eq:rdp}
	&D_{\alpha} (\mathcal{M}(D)||\mathcal{M}(D')) \nonumber \\
	:=& \frac{1}{\alpha - 1 } \log \expected{(\frac{Pr[\mathcal{M}(D)=\delta]}{Pr[\mathcal{M}(D')=\delta]})^{\alpha}}{\delta \sim \mathcal{M}(D')} \leq \varepsilon
\end{align}
\end{definition}

There are two key properties of DP algorithms that will be useful in our \emph{composition} and \emph{post-processing}. Below we briefly review these two properties specifically for the widely-used R\'enyi-DP definition, but they apply to all the DP algorithms.

\begin{lemma}[Adaptive Composition of RDP~\cite{mironov2017renyi}]\label{lem:composition}
	Consider two randomized mechanisms  $\mathcal{M}_1$ and $\mathcal{M}_2$ that provide  $(\alpha,\varepsilon_1)$-RDP and  $(\alpha,\varepsilon_2)$-RDP, respectively. Composing $\mathcal{M}_1$ and $\mathcal{M}_2$ results in a mechanism with 
 $(\alpha,\varepsilon_1 + \varepsilon_2)$-RDP.
\end{lemma}

\begin{lemma}[Post-processing of RDP~\cite{mironov2017renyi}]
Given a randomized mechanism  that is $(\alpha,\varepsilon)$-RDP, applying a randomized mapping function on it does not increase its privacy budget, i.e., it will result in another $(\alpha,\varepsilon)$-RDP mechanism.
\label{lemma:ec_post}
\end{lemma}

\subsubsection{Differentially Private ML Algorithms We Use}
Several works have  used differential privacy in  traditional  machine learning to protect the privacy of the training data~\cite{li2014privacy,chaudhuri2011differentially,feldman2018privacy,zhang2016differential,bassily2014private}.  We use two of the commonly-used algorithms for DP deep learning: DP-SGD~\cite{abadi2016deep}, and DP-FTRL~\cite{kairouz2021practical}.  At a high level, to update the model in each training round, DP-SGD first samples a minibatch of examples uniformly at random, clips the gradient of each example to limit the sensitivity of a gradient update,  and then adds independent Gaussian noise to gradients that is calibrated to achieve the desired DP guarantee.  In contrast, in each training round, DP-FTRL takes a minibatch of examples (no requirement of sampling), clips each example's gradient to limit sensitivity, and adds correlated Gaussian noise calibrated to achieve the desired DP guarantee.
\section{Using Checkpoint Aggregates to Improve Accuracy of Differentially Private ML}\label{acc}

In this section, we first detail our novel and general  \emph{adaptive aggregation training framework} that leverages past checkpoints (recall a checkpoint is just an intermediate model iterate $\theta_t$) during training, and provide two instantiations of it. 
We also design four checkpoint aggregation methods that can be used for inference over a given sequence of checkpoints.
Finally, we provide a theoretical analysis for improved privacy-utility trade-offs due to some of the checkpoint aggregations.

\mypar{Why can we post-process intermediate DP ML checkpoints?}
Before delving into the details of our checkpoints aggregation methods, it is useful to note that  the privacy analyses for the DP algorithms we consider in this paper, i.e., DP-SGD~\cite{abadi2016deep} and DP-FTRL~\cite{kairouz2021practical}, use the adaptive composition (Lemma~\ref{lem:composition}) across training rounds. This implies that all the intermediate checkpoints are also DP, which allows us to release of all intermediate checkpoints computed during training. Furthermore, as all checkpoints are DP, due to the post-processing property of DP (Lemma~\ref{lemma:ec_post}), one can process/use these checkpoints without incurring additional privacy cost.

\vspace*{-.25em}
\subsection{Using Checkpoint Aggregations for Training}\label{acc:tr_framework}

Algorithm~\ref{alg:atf} describes our general adaptive aggregation training framework.
Apart from the parameters needed to run the DP algorithm $\calA$, it uses a \emph{checkpoint aggregation} function $f_{\sf AGG}$ to compute an aggregate checkpoint $\theta^{\sf AGG}_{t+1}$ from the checkpoints $(\theta_{t+1}, \theta_{t}, \ldots , \theta_0)$ at each step $t$. Consequently, $\calA$ uses $\theta^{\sf AGG}_{t+1}$ for its next training step.
Note that Algorithm~\ref{alg:atf} has two hyperparameters: (1) $\tau$ that decides when to start training over the past checkpoints aggregate, and (2) parameter $p$ specific to $f_{\sf AGG}$ which we detail below, along with $f_{\sf AGG}$s. 
Due to the post-processing property of DP, using $f_{\sf AGG}$ does not incur any additional privacy cost. 
Though our framework can incorporate any custom $f_{\sf AGG}$, we present two natural instantiations for $f_{\sf AGG}$ and extensively evaluate them.

\begin{algorithm}
   \caption{Our {adaptive aggregation training framework}.}
   \label{alg:atf}
\begin{algorithmic}
\STATE {\bfseries Input:} Iterative DP ML algorithm $\calA$, private dataset $D$, initial model $\theta_0$, number of training steps $T$, checkpoints aggregation function $f_{\sf AGG}$ and its parameter $p$ (EMA coefficient $\beta$ for \algtr{EMA} and number of last $k$ checkpoints for \algtr{UTA}), the step to start training over past aggregate $\tau$
\STATE $\theta^{\sf AGG}_0 = \theta_0$.
\FOR{$t=0$ {\bfseries to} $T$}
\IF{$t\geq\tau$}
    \STATE $\theta_{t+1} \leftarrow \calA(\theta^{\sf AGG}_t; D)$.
    \STATE $\theta^{\sf AGG}_{t+1} = f_{\sf AGG}(\{\theta_{t+1}, \theta_{t}, \ldots , \theta_0\}, p)$.
\ELSE
    \STATE $\theta_{t+1} \leftarrow \calA(\theta_t; D)$.
\ENDIF
\ENDFOR
\STATE \textbf{Return} $\theta^{\sf AGG}_{t+1}$
\end{algorithmic}
\end{algorithm}

\mypar{\em Exponential Moving Average (EMA)}
Our first proposal uses an EMA function to aggregate all the past checkpoints at training step $t$. 
Starting from the latest checkpoint, EMA assigns exponentially decaying weights to each of the previous checkpoints.
At step $t$, EMA maintains a moving average $\theta^{\sf EMA}_t$ that is a weighted average of $\theta^{\sf EMA}_{t-1}$ and the latest checkpoint, $\theta_t$ . 
This is formalized as follows: 
\begin{align}
    \theta^{\sf EMA}_t = (1 - \beta_t) \cdot \theta^{\sf EMA}_{t-1} + \beta_t \cdot \theta_{t} \label{eqn:ema}
\end{align}

\mypar{\em Uniform Tail Averaging (UTA)}
Our second proposal uses a UTA function to aggregate past $k$ checkpoints. Specifically, for step $t$, UTA computes the \emph{parameter-wise} mean of the past $\min\{t+1,k\}$ checkpoints. We formalize this as:
\begin{align}
    \theta^{\sf UTA}_{t} = \frac{1}{\min\{ t+1, k\}} \sum^{t}_{i=\max\{0, t-(k-1)\}} \theta_{i} \label{eqn:uta}
\end{align}

\subsection{Using Checkpoint Aggregations for Inference}\label{acc:inf_framework}

In many scenarios, e.g., where a DP ML technique has been applied to release a sequence of checkpoints, checkpoint aggregation functions can be used as post-processing functions over the released checkpoints to reduce bias of the technique at inference time.
In this section, we design various aggregation methods towards this goal.

We note that~\citep{tanle19, BrockDSS21} have used EMA (Equation~\ref{eqn:ema}) to improve the performance of ML techniques at inference time in non-private settings. \citet{de2022unlocking} extend EMA to DP-SGD, but use EMA coefficients $\beta$ suggested from non-private settings; we denote this EMA baseline by $\emaBase$. However, as we will show in Section~\ref{sec:empEval}, even a coarse-grained tuning of $\beta$ provides significant accuracy gains in DP settings.
To highlight the crucial difference with the instantiation in Section~\ref{acc:tr_framework}, we use \algtr{EMA} to denote when we use aggregation adaptively in training (Algorithm~\ref{alg:atf}), and \alginf{EMA} to denote when we use the aggregation only for inference. Since UTA (Equation~\ref{eqn:uta}) can be applied as an aggregation at inference time, we similarly define \algtr{UTA} and \alginf{UTA}.

\mypar{Outputs aggregation functions}
So far, our aggregation functions have focused on aggregating parameters of intermediate checkpoints. Next, we design two aggregation functions that, given a sequence of checkpoints $\theta_i, i \in [t]$, compute a function of the outputs of the checkpoints and use it for making predictions.

\mypar{\em Output Predictions Averaging (OPA)}
For a given test sample $\mathbf{x}$, OPA first computes prediction vectors $f_{\theta_{i}}(\mathbf{x})$ of the last $k$ checkpoints, i.e., checkpoints from steps $\in [t-(k-1), t]$, averages the prediction vectors, and computes argmax of the average vector as the final output label. We formalize OPA as follows:
\begin{align}
    \hat{y}_\text{opa}(\mathbf{x}) = \text{argmax} \Big(\frac{1}{k} \sum^{t}_{i=t-(k-1)} f_{\theta_{i}}(\mathbf{x}) \Big)
\end{align}

\mypar{\em Output Labels Majority Vote (OMV)}  For a given test sample $\mathbf{x}$, OMV computes output prediction labels, i.e., $\text{argmax}\ f_{\theta_{i}}(\mathbf{x})$ for the last $k$ checkpoints. 
Finally, it outputs the majority label among the $k$ labels (breaking ties arbitrarily) for inference. 
We formalize OMV as follows:
\begin{align}
    \hat{y}_\text{omv}(\mathbf{x}) =  \text{Majority} \big( \text{argmax}( f_{\theta_{i}}(\mathbf{x}) ) ^t_{i=t-(k-1)}\big)
\end{align}

\subsubsection{Improved Excess Risk via Tail Averaging}
\label{acc:theory}

Results from \cite{shamir2013stochastic} can be used to demonstrate how a family of checkpoint aggregations, which includes \alginf{UTA} (Section~\ref{acc:inf_framework}), provably improves the privacy/utility trade-offs compared to that of the last checkpoint of DP-(S)GD. To formalize the problem, we define the following notation: Consider a data set $D=\{d_1,\ldots,d_n\}$ and a loss function $\calL(\theta;D)=\frac{1}{n}\sum\limits_{i=1}^n\ell(\theta;d_i)$, where each of the loss function $\ell$ is convex and $L$-Lipschitz in the first parameter, and $\theta\in\calC$ with $\calC\subseteq\mathbb{R}^p$ being a convex constraint set. 
We analyze the following variant of DP-GD (Algorithm~\ref{alg:DPSGD}), which is guaranteed to be $\rho$-zCDP defined below. Note that using~\cite{bun2016concentrated}, it is easy to convert the privacy guarantee to an $(\epsilon,\delta)$-DP guarantee. Moreover, while our analytical result is for DP-GD (due to brevity), it extends to DP-SGD with mild modifications to the
proof.

\begin{defn}[zCDP~\cite{bun2016concentrated}]
A randomized algorithm $M : \mathcal{D}^* \to \mathcal{Y}$ is $\rho$-zero-concentrated differentially private (zCDP) if, for all neighbouring datasets $D,D' \in \mathcal{D}^*$ (i.e., datasets differing in one data sample) and all $\alpha \in (1, \infty)$, we have
$$ {\sf D}_\alpha \left(M(D) \| M(D')\right) \leq \rho \alpha $$
where ${\sf D}_\alpha \left(M(D) \| M(D')\right)$ is the $\alpha$-R\'enyi divergence between the distribution of $M(D)$ and $M(D')$.
\label{defn:zcdp}
\end{defn}

\begin{algorithm}[tb]
   \caption{DP Gradient Descent (DP-GD)}
   \label{alg:DPSGD}
\begin{algorithmic}
   \STATE $\theta_0\leftarrow \mathbf{0}^p$.
   \FOR{$t\in[T]$}
        \STATE $\theta_{t+1}\leftarrow\Pi_{\calC}\left(\theta_t-\eta_t\left(\nabla\calL(\theta_t;D)+b_t\right)\right)$, where $b_t\sim\calN\left(0,\frac{L^2T}{2n\rho}\mathbb{I}_{p\times p}\right)$, and $\Pi_\calC\left(\cdot\right)$ being the $\ell_2$-projection onto the set $\calC$.
   \ENDFOR
\end{algorithmic}
\end{algorithm}

We will provide the utility guarantee for this algorithm by directly appealing to the result of~\cite{shamir2013stochastic}. 
For a given $\alpha\in (0,1)$, \alginf{UTA} corresponds to the average of the last $\alpha T$ models, i.e., 
\begin{align}
    \theta^{\sf UTA}_t=\frac{1}{\alpha T}\sum\limits_{t=(1-\alpha)T+1}^T\theta_t
\end{align}
One can also consider~\emph{polynomial-decay averaging} (PDA) with parameter $\gamma \geq 0$, defined as follows:
\begin{align}
    \theta^{\sf PDA}_t=\left(1-\frac{\gamma+1}{t+\gamma}\right)\theta^{\sf PDA}_{t-1}+\frac{\gamma+1}{t+\gamma}\cdot\theta_t
\end{align}

For $\gamma = 0$, PDA matches \alginf{UTA} over all iterates. As $\gamma$ increases, PDA places more weight on later iterates; in particular, if $\gamma = cT$, the averaging is  similar to \alginf{EMA} (Section~\ref{acc:inf_framework}), since as $t \rightarrow T$ the decay parameter $\frac{\gamma+1}{t+\gamma}$ approaches a constant $\frac{c}{c+1}$. In that sense, PDA can be viewed as a method interpolating between \alginf{UTA} and \alginf{EMA}. From \cite{shamir2013stochastic}, we can derive the following bounds on the different methods:

\begin{thm}
There exists a choice of learning rate $\eta_t$ and the number of time steps $T$ in DP-GD (Algorithm~\ref{alg:DPSGD}) such that the following hold for $\alpha=\Theta(1)$:
$$\mathbb{E}\left[\calL\left(\privTa;D\right)\right]-\min\limits_{\theta\in\calC}\calL(\theta;D)=\calO\left(\frac{L\ltwo{\calC}\sqrt{p}}{n\rho}\right)$$
and
$$\mathbb{E}\left[\calL(\theta_T;D)\right]-\min\limits_{\theta\in\calC}\calL(\theta;D)=\calO\left(\frac{L\ltwo{\calC}\sqrt{p}\log(n)}{n\rho}\right).$$

Furthermore, for $\gamma=\Theta(1)$, we have,
$$\mathbb{E}\left[\calL\left(\theta^{\sf PDA}_T;D\right)\right]-\min\limits_{\theta\in\calC}\calL(\theta;D)=\calO\left(\frac{L\ltwo{\calC}\sqrt{p}}{n\rho}\right).$$
\label{thm:tail}
\end{thm}
\begin{proof}
These bounds build on Theorems 2 and 4  of~\cite{shamir2013stochastic}. If we choose $T = \lceil n \rho\rceil$ and set $\eta_t$ appropriately, the proof of Theorem 2 \citep{shamir2013stochastic} implies the following for $\theta^{\sf UTA}_T$:
$$\mathbb{E}\left[\calL\left(\theta^{\sf UTA}_T;D\right)\right]-\min\limits_{\theta\in\calC}\calL(\theta;D) = O\left(\frac{L \ltwo{\calC} \sqrt{p}}{n \rho} \log\left(\frac{1}{\alpha}\right)\right).$$

Setting $\alpha = \Theta(1)$ gives the theorem's first part, and $\alpha T = 1$, i.e., $1/\alpha = T = \lceil n \rho \rceil$ gives the second.
The third follows from modifying Theorem 4 of \cite{shamir2013stochastic} for the convex case (see the end of Section 4 of \cite{shamir2013stochastic} for details).
\end{proof}

\textbf{Theorem~\ref{thm:tail} implies that the excess empirical risk for $\theta_T$ is higher by factor of $\log(n)$ in comparison to $\theta^{\sf UTA}_T$ and $\theta^{\sf PDA}_T$}. {For step size selections typically used in practice (e.g., fixed or inverse polynomial step sizes), the last iterate will suffer from the extra $\log(n)$ factor, and we do not know how to avoid it. 
Furthermore, \citet{Harvey2019TightAF} showed that this is unavoidable in the non-private, high probability regime.
{\citet{Jain2021LastIterate} show that for carefully chosen step sizes, the logarithmic factor can be removed, and \citet{feldman2019private} extend this analysis to a  DP-SGD variant with varying batch sizes. Unlike those methods, averaging can be done as  post-processing of DP-SGD outputs, rather than a modification of the algorithm.}}
\section{Empirical Evaluation}
\label{sec:empEval}
In this section, we first describe experimental setup, followed by experiments in a user-level and sample-level DP settings.

\subsection{Experimental Setup}\label{sec:empEval:setup}

\subsubsection{Datasets and ML Settings}\label{sec:empEval:setup:dataset}
We evaluate our checkpoints aggregation algorithms on three benchmark datasets (StackOverflow, CIFAR10, CIFAR100) and one proprietary production-grade dataset (pCVR) in two different settings.

\mypar{StackOverflow}
StackOverflow~\cite{stackoverflow} is a natural-language dataset containing questions and answers from StackOverflow forum. We use it to train a model for next word prediction task. StackOverflow is a \emph{user-keyed dataset}, i.e., all the samples in the data are owned by some users. It is a large dataset containing training data of total of 342,477 users and over 135M samples. The original test data contains data of 204,088 users; following~\cite{reddi2020adaptive}, we sample 10,000 users for validation data. Following~\cite{reddi2020adaptive}, we use vocabulary of top-10,000 words from StackOverflow data.

We use simulated federated learning (FL)~\cite{mcmahan2017communication} to train on StackOverflow data. In each FL round, a central server (model trainer) broadcasts a global model to all users, users share gradient updates that they compute using the model and their local dataset. The central server then aggregates all user updates and updates the global model to be used for the following FL rounds.

\mypar{CIFAR Datasets}
We experiment with CIFAR10 and CIFAR100 datasets. CIFAR10 (CIFAR100)~\cite{krizhevsky2009learning} is a 10-class (100-class) image classification task and contains 60,000 $32\times32$ color~(RGB) images~(50,000 images as training set and 10,000 images as test set). We use centralized ML for CIFAR10 (CIFAR100) training, i.e., when model trainer collects all data in one place and trains a model on it.

\mypar{pCVR (Predicted Conversion Rate) Dataset} 
This is a proprietary, production-grade dataset (also used in~\cite{chua2024training, denison2022private}), where each example corresponds to an ad click, and the task is to predict
whether a conversion takes place after the click, which is commonly referred as predicted conversion rate (pCVR). 
As users' clicking and conversion information is highly sensitive, such data needs to be protected with differential privacy.
We use centralized ML for training, similar to CIFAR datasets. This dataset contains significantly more examples, by orders of magnitude, than the aforementioned datasets.

\begin{table}
\caption{StackOverflow LSTM architecture details.}
\label{tab:so_arch}
\centering
\begin{tabular} {|c|c|c|}
    \hline
    Layer & Output shape & Parameters \\ \hline
    Input & 20 & 0 \\
    Embedding & (20, 96) & 960384  \\
    LSTM & (20, 670) & 2055560 \\
    Dense & (20, 96) & 64416 \\
    Dense & (20, 10004) & 970388 \\
    Softmax & - & - \\
    \hline
\end{tabular}
\end{table}

\begin{figure}[ht]
\hspace*{-.5em}
\centering
\includegraphics[scale=.45]{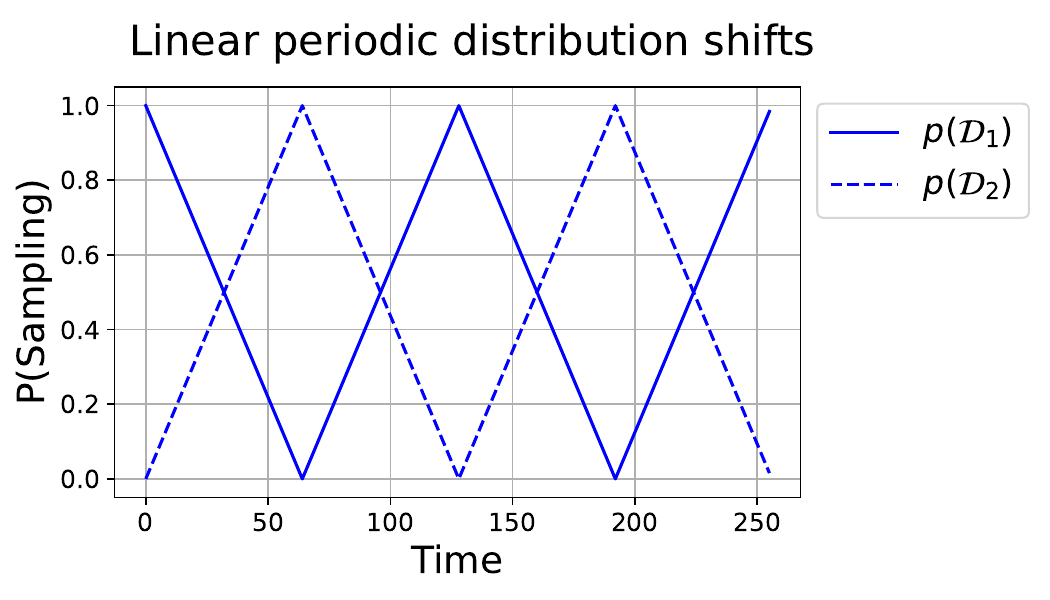}
\vspace*{-.5cm}
\caption{Probability of sampling users or samples from two periodically shifting distributions $\mathcal{D}_{\{1,2\}}$.}
\label{fig:linear_pds}
\end{figure}

\subsubsection{Periodic Distribution Shift (PDS) Settings}
\label{acc:exp:pds}
The distribution of data sampled from the datasets discussed above is almost uniform throughout the training; we call such datasets \textbf{\em original datasets}.  However, in many real-world settings, e.g., in FL, the training data distribution may vary over time. \citet{zhu2021diurnal} demonstrate the adverse impacts of distribution shifts in training data on the performances of resulting FL models. Due to their practical significance, we consider settings where the training data distribution models \emph{diurnal} variations, i.e., it is a function of two oscillating distributions (see Figure~\ref{fig:linear_pds} for an example).  Such a scenario commonly occurs in FL training, e.g., when a model is trained with client devices participating from two significantly different time zones.

Following~\cite{zhu2021diurnal}, we consider a setting where training data is a combination of clients/samples drawn from two disjoint data distributions which oscillate over time (Figure~\ref{fig:linear_pds}). Here, the probabilities of sampling at time $t$ are: $p(\mathcal{D}_1, t) = \big| 2\frac{t\ \text{mod}\ T}{T} - 1\big|$ $p(\mathcal{D}_2, t) = (1 - p(\mathcal{D}_1, t))$, where $T$ is the period of oscillation of $\mathcal{D}_{\{1,2\}}$.

\mypar{Simulating periodic distribution shifting settings}
To simulate such periodically shifting distribution for StackOverflow, we use $\mathcal{D}_1$  with only questions and $\mathcal{D}_2$ with only answers from users. Then, we draw clients from $\mathcal{D}_{\{1,2\}}$. Apart from data distribution, the rest of experimental setup is the same as before. We use test and validation data same as for the original StackOverflow setting.
To simulate PDS CIFAR10/CIFAR100, we use $\mathcal{D}_{\{1,2\}}$ such that $\mathcal{D}_1$ and  $\mathcal{D}_2$ respectively contain the data from even and odd classes of the original data; the rest of the sampling strategy is the same as described in Section~\ref{acc:exp:pds}.

\subsubsection{Model Architectures and Training Details}\label{appdx:training_hparam}
Below we detail the model architectures, DP ML algorithms, and various hyperparameters we use to obtain our results. 

Note that, for each of the tasks we evaluate, we select the state-of-the-art DP ML algorithm as the baseline algorithm and demonstrate improvements on top of the performances of such state-of-the-art DP ML algorithms. For instance, we use DP-FTRL for StackOverflow task as it provides state-of-the-art performance on StackOverflow; DP-SGD does not perform well on StackOverflow hence we omit it from StackOverflow experiments. For the same reason, we use DP-SGD for the rest of the tasks.

\mypar{StackOverflow training}
For StackOverflow, we follow the state-of-the-art DP training in \citep{kairouz2021practical,denisov2022improved} and train a one-layer LSTM using DP-FTRL with momentum in Tensorflow Federated framework~\citep{abadi2016tensorflow} for $\eps \in \{8.2, 18.9\}$, which corresponds to $\rho$-zCDP with $\rho \in \{1.08, 4.31\}$, respectively. We process 100 users in each FL round and train for total of 2,000 rounds. For experiments with DP, we fix the privacy parameter $\delta$ to $10^{-6}$ for StackOverflow ensuring that $\delta < n^{-1}$, where $n$ is the number of users in StackOverflow. Since StackOverflow data is naturally keyed by users, the privacy guarantees here are at user-level, in contrast to the example-level privacy for CIFAR10.

Tables~\ref{tab:so_hparams} and~\ref{tab:pds_so_hparams} provide the hyperparameters we use for training aggregations (\algtr{UTA}, \algtr{EMA}) using DP-FTRL.

\mypar{CIFAR10 training}
Following the setup of the state-of-the-art DP-SGD training in \citep{de2022unlocking}, we train a WideResNet-16-4 with depth 16 and width 4 using DP-SGD~\citep{DP-DL} in JAXline~\citep{deepmind2020jax} for $\eps \in \{1, 8\}$. We fix clip norm to 1, batch size to 4096 and augmentation multiplicity to 16 as in~\citep{de2022unlocking}. For experiments with DP, we fix the privacy parameter $\delta$ to $10^{-5}$ on CIFAR10 ensuring that $\delta < n^{-1}$, where $n$ is the number of examples in CIFAR10. Here the DP guarantee is at sample-level.

For training on CIFAR10, we use the state-of-the-art DP-SGD parameters from~\cite{de2022unlocking} as follows: we set learning rate and noise multiplier, respectively, to 2 and 10 for $\epsilon=1$ and to 4 and 3 for $\epsilon=8$. We stop the training when the intended privacy budget exhausts. All the hyperparameters we use to generate the results of Table~\ref{tab:cifar10} are in Table~\ref{tab:c10_hparams}. 

\mypar{CIFAR100 training}
Similarly to~\cite{de2022unlocking}, for CIFAR100, we use Jaxline~\citep{jax2018github} and use DP-SGD to \emph{fine-tune the last, classifier layer of} a WideResNet with depth 28 and width 10 that is pre-trained on entire ImageNet data. We fix clip norm to 1, batch size to 16,384 and augmentation multiplicity to 16. Then, we set learning rate and noise multiplier, respectively, to 3.5 and 21.1 for $\epsilon=1$ and to 4 and 9.4 for $\epsilon=8$. For periodic distribution shifting (PDS) CIFAR100, we set learning rate and noise multiplier, respectively, to 4 and 21.1 for $\epsilon=1$ and to 5 and 9.4 for $\epsilon=8$. We stop the training when privacy budget exhausts. Setup for training aggregations is the same as for CIFAR10 above; hyperparameters used to generate results in Table~\ref{tab:cifar100} are in Table~\ref{tab:c100_hparams}.

\mypar{pCVR Training}
We employ a multi-encoder model architecture, where each encoder is responsible for encoding a specific class of features (e.g., ads features).
We consider sample level privacy with  $\epsilon=6$ and $\delta=\frac{1}{n}$, where $n$ is the number of examples, as these are the parameters that are of production requirement.

The model is trained with logistic loss and is measured by the test \textit{AUC loss} (i.e., 1 - AUC), as is commonly done for pCVR tasks \citep{denison2022private,chua2024training}.
In real-world advertising scenarios, the pCVR models' outputs (i.e., the predicted conversion probability) are often passed directly to downstream models for calculating final ad bids, instead of being converted to binary predictions. Therefore, we use AUC-loss instead of other commonly used classification metrics, such as accuracy. For the same reason, Majority voting (\algtr{OMV}) is not applicable for this task.

We adopt a two-stage hyperparameter-tuning strategy for DP-SGD. We first tune the batch size, number of steps, clip norm, and learning rate for baseline DP-SGD, and then, with the above fixed, tune the hyperparameters in Section \ref{appdx:hparam_tuning}. This is done primarily due to the significant training cost associated with pCVR.

\begin{table}
\caption{Tuning the EMA coefficient can provide significant gains in accuracy over the default value of 0.9999 from \cite{de2022unlocking} implying the need to tune EMA coefficients for each different privacy budget to achieve the best performances. Results below are for original CIFAR10 dataset.}
\vspace*{-.3cm}
\label{tab:cifar10_ema}
\centering
\begin{tabular} {|c|c|c|c|c|}
    \hline
    \multirow{2}{*}{Privacy level} & \multicolumn{4}{c|}{EMA coefficient} \\ \cline{2-5}
    & 0.9 & 0.95 & 0.99 & 0.999 (\cite{de2022unlocking}) \\ \hline
    $\epsilon = 8$ & 79.41 & 79.35 & \bf 79.41 & 79.16 \\ \hline
    $\epsilon = 1$ & 56.59 & \bf 56.61 & 56.06 & 56.05 \\ \hline
\end{tabular}
\vspace*{-.2cm}
\end{table}

\begin{algorithm}
   \caption{Hyperparameter tuning for training aggregations.}
   \label{alg:ht_agg_tr}
\begin{algorithmic}
   \STATE {\bfseries Input:} Adaptive training algorithm $\calA^{\sf Ada}$ (Algorithm~\ref{alg:atf}) with aggregation function $f_{\sf AGG}$ and its hyperparameter $p$, range of hyperparameters $\{p, \tau\}$ for grid search $R_{p,\tau}$, validation set $D_v$, $T$ training steps, Initial $\theta_0$.
   \STATE \textbf{Initialize}: ${\sf Acc}_{max}\leftarrow0$, $\theta_{best}\leftarrow\theta_{0}$, $\{p_{best},\tau_{best}\}\leftarrow\{1,0\}$.
   \FOR{$\{p, \tau\}$ in $R_{p,\tau}$}
   \STATE Run $\calA^{\sf Ada}$ for $T$ steps with $f_{\sf AGG}$, $p$, $\tau$ as detailed in Algorithm~\ref{alg:atf}
   \STATE $$\theta^{\sf Ada}_T\leftarrow \calA^{\sf Ada}(f_{\sf AGG}, p, \tau, \theta_0)$$
   \STATE Compute accuracy of the output checkpoint on validation set: 
   \STATE $${\sf Acc}_{\sf Ada} = {\sf Acc}(\theta^{\sf Ada}_T, D_v)$$
   \IF{${\sf Acc}_{\sf Ada}>{\sf Acc}_{max}$}
       \STATE ${\sf Acc}_{max}\leftarrow{\sf Acc}_{\sf AGG}$, $\theta_{best}\leftarrow\theta^{\sf Ada}_{T}$, $\{p_{best},\tau_{best}\leftarrow\{p,\tau\}$
    \ENDIF
   \ENDFOR
   \STATE \textbf{Return} $\theta_{best}$, $p_{best}$, $\tau_{best}$
\end{algorithmic}
\end{algorithm}

\subsubsection{Hyperparameters Tuning for Our Aggregations}\label{appdx:hparam_tuning}
Performances of our training and inference aggregations (Section~\ref{acc:tr_framework},~\ref{acc:inf_framework}) depend heavily on certain hyperparameters; we first discuss advantages and disadvantages of these hyperparameters' values. In \alginf{EMA} and \algtr{EMA}, EMA coefficient $\beta$ sets the weights of the checkpoints. Specifically larger $\beta$ gives higher weight to newer checkpoints which are generally better than previous checkpoints hence we tune $\beta$ starting from 0.5.
The number $k$ of past checkpoints aggregated affects the performances of the rest of the training and inference aggregations. Very large $k$ includes contribution of checkpoints from early training while very small $k$ may ignore good checkpoints, both of which may hurt the performance of the final aggregate. Therefore, we tune $k$ in a fairly wide range starting from $k=3$ up to $k=200$.
Next, we detail the empirical methodology we follow to obtain the best hyperparameters for our aggregations. 

\mypar{Training aggregations}
Our use a simple grid-search strategy to tune hyperparameters as detailed in Algorithm~\ref{alg:ht_agg_tr}. Note that there are two hyperparameters to tune: aggregation parameters $p$ and step to start training over past aggregate $\tau$. For \algtr{EMA}, $p$ in Algorithm~\ref{alg:ht_agg_tr} is the EMA coefficient $\beta$ in~\eqref{eqn:ema}, and we tune $\beta\in\{0.5, 0.6, 0.7, 0.8, 0.85, 0.9, 0.95, 0.99, 0.999, 0.9999\}$ for all datasets. For StackOverflow we fix $\tau=0$ while for CIFAR10 we tune $\tau\in\{100, 200,\ldots \tau^*\}$ where $\tau^*$ is largest multiple of 100 smaller than total number of steps $T$; for CIFAR100 we tune $\tau \in \{50, 100,\ldots, 250\}$. For \algtr{UTA}, $p$ in Algorithm~\ref{alg:ht_agg_tr} is the number of $k$ past checkpoints to aggregate. For CIFAR10/CIFAR100 we tune $k \in \{2, 3, 5, 10, 20,..., 200\}$, for pCVR we tune $k\in\{3, 5\}$ and for StackOverflow we tune $k \in \{2, 3, 5, 10, 20,..., 200\}$ for \algtr{UPA}. 
Finally note that, in case of StackOverflow, we use inference aggregation after producing all intermediate checkpoints using training aggregations. So we follow the hyperparameter tuning strategies for training and inference aggregations in sequence.

\begin{algorithm}
   \caption{Hyperparameter tuning for inference aggregations.}
   \label{alg:ht_agg_inf}
\begin{algorithmic}
   \STATE {\bfseries Input:} Intermediate checkpoints $(\theta_{T-1}, \ldots \theta_0)$ from $T$ training steps, checkpoints aggregation function $f_{\sf AGG}$ and its hyperparameter $p$, range of $p$ for grid search $R_p$, validation set $D_v$.
   \STATE \textbf{Initialize}: ${\sf Acc}_{max}\leftarrow0$, $\theta_{best}\leftarrow\theta_{T-1}$, $p_{best}\leftarrow1$.
   \FOR{$p$ in $R_p$}
   \STATE Compute aggregated checkpoint
   \STATE $$\theta^{\sf AGG}_{T} = f_{\sf AGG}(\{\theta_{T-1}, \ldots \theta_0\}, p)$$
   \STATE Compute accuracy of aggregated checkpoint on validation set: 
   \STATE $${\sf Acc}_{\sf AGG} = {\sf Acc}(\theta^{\sf AGG}_{T}, D_v)$$
   \IF{${\sf Acc}_{\sf AGG}>{\sf Acc}_{max}$}
       \STATE ${\sf Acc}_{max}\leftarrow{\sf Acc}_{\sf AGG}$, $\theta_{best}\leftarrow\theta^{\sf AGG}_{T}$, $p_{best}\leftarrow p$
    \ENDIF
   \ENDFOR
   \STATE \textbf{Return} $\theta_{best}$, $p_{best}$
\end{algorithmic}
\end{algorithm}

\mypar{Inference aggregations}
Our simple grid-search strategy to tune hyperparameters is detailed in Algorithm~\ref{alg:ht_agg_inf}.
For \alginf{EMA}, $p$ in Algorithm~\ref{alg:ht_agg_inf} is the EMA coefficient $\beta$ in~\eqref{eqn:ema}.
\citet{de2022unlocking} simply use $\beta$ that works the best in non-private settings. However, tuning $\beta\in\{0.85, 0.9, 0.95, 0.99, 0.999, 0.9999\}$, we observe that the best $\beta$ for private and non-private settings need not be the same (Table~\ref{tab:cifar10_ema}). For instance, for CIFAR10, for $\epsilon$ of 1 and 8, \alginf{EMA} coefficient of 0.95 and 0.99 perform the best and outperform 0.9999 by 0.6\% and 0.3\%, respectively. Hence, we advise future works to perform tuning of EMA coefficient. Full results are given in Table~\ref{tab:cifar10_ema}. For \alginf{UTA}, OPA and OMV, $p$ in Algorithm~\ref{alg:ht_agg_inf} is the number of last checkpoints $k$ to aggregate. We tune $k$ in the same range as in training aggregations.

\begin{table*}[ht]
\caption{Test accuracy gains due to checkpoints aggregations for original and PDS StackOverflow. We present techniques from prior works (DP-FTRL baseline~\cite{kairouz2021practical, denisov2022improved}, and ${\sf EMA}_{\sf baseline}$~\cite{de2022unlocking}) in .}
\label{tab:stackoverflow}
\centering
\resizebox{\linewidth}{!}{%
\begin{tabular} {|c|c|c|c|c|c|c|c|c|}
    \hline
    {DP} &  & \multicolumn{2}{c|}{Training Aggregations} & \multicolumn{5}{c|}{Inference Aggregations} \\ \cline{3-9}
    
    $(\eps)$ &  & ${\sf EMA}_{\sf tr}$ & ${\sf UTA}_{\sf tr}$ & & {${\sf EMA}_{\sf inf}$} & ${\sf UTA}_{\sf inf}$ & OPA & OMV \\ \hline
    \multicolumn{9}{|c|}{StackOverflow; DP-FTRL; user-level privacy} \\ \hline
    $\infty$ &  & 25.72 \tiny{$\pm$ 0.02} & \bf 25.98 \tiny{$\pm$ 0.01} &  & {25.79 \tiny{$\pm$ 0.01}} & 25.81 \tiny{$\pm$ 0.02} & 25.79 \tiny{$\pm$ 0.01} & 25.78 \tiny{$\pm$ 0.01} \\ \hline
    18.9 &  & 23.56 \tiny{$\pm$ 0.02} & \bf 23.90 \tiny{$\pm$ 0.02} &  & {23.63 \tiny{$\pm$ 0.01}} & 23.84 \tiny{$\pm$ 0.01} & 23.60 \tiny{$\pm$ 0.02} & 23.57 \tiny{$\pm$ 0.02} \\ \hline
    8.2 &  & 22.43 \tiny{$\pm$ 0.04} & \bf 22.74 \tiny{$\pm$ 0.04} &  & {22.54 \tiny{$\pm$ 0.02}} & 22.70 \tiny{$\pm$ 0.03} & 22.57 \tiny{$\pm$ 0.04} & 22.52 \tiny{$\pm$ 0.04} \\ \hline
    \multicolumn{9}{|c|}{\emph{Periodic Distribution Shifting (PDS)} StackOverflow; DP-FTRL; user-level privacy} \\ \hline
    $\infty$ &  & 23.97 \tiny{$\pm$ 0.04} & \bf 24.26 \tiny{$\pm$ 0.02} &  & {23.92 \tiny{$\pm$ 0.12}} & 23.98 \tiny{$\pm$ 0.02} & 23.87 \tiny{$\pm$ 0.01} & 23.91 \tiny{$\pm$ 0.07} \\ \hline
    $18.9$ &  & 21.90 \tiny{$\pm$ 0.04} & \bf 22.17 \tiny{$\pm$ 0.03} &  & {21.82 \tiny{$\pm$ 0.07}} & 22.04 \tiny{$\pm$ 0.11} & 21.99 \tiny{$\pm$ 0.13} & 21.95 \tiny{$\pm$ 0.16} \\ \hline
    $8.2$ &  & 20.37 \tiny{$\pm$ 0.06} & \bf 20.81 \tiny{$\pm$ 0.05} &  & {20.36 \tiny{$\pm$ 0.06}} & 20.75 \tiny{$\pm$ 0.05} & 20.67 \tiny{$\pm$ 0.03} & 20.72 \tiny{$\pm$ 0.16} \\ \hline
\end{tabular}}
\vspace*{-.5em}
\end{table*}

\begin{figure*}[ht]
\centering
\hspace*{-.8cm}
\begin{subfigure}{\linewidth}
  \centering
  \includegraphics[width=\linewidth, height=.25\linewidth]{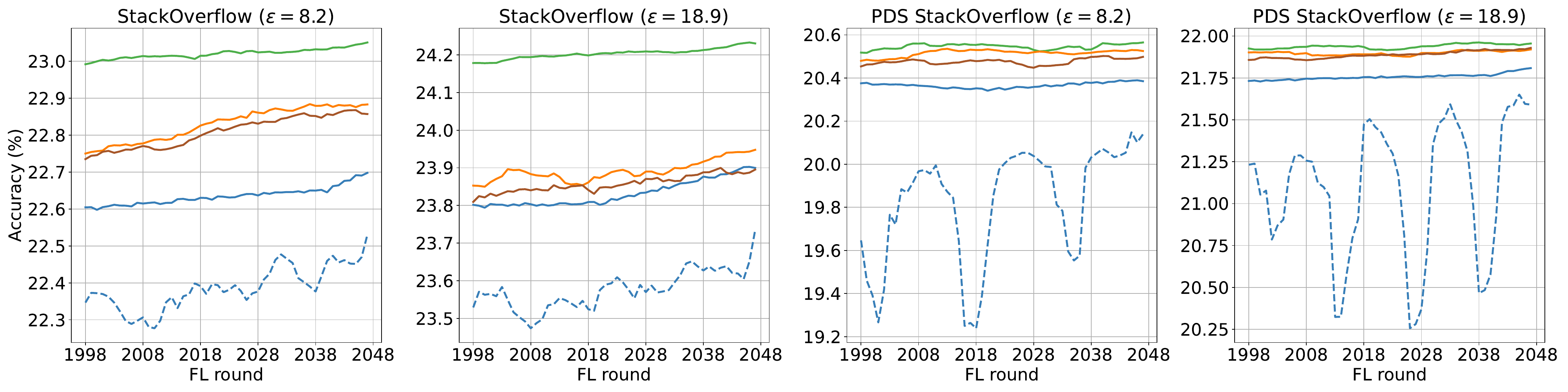}
\end{subfigure}%
\vspace*{-.01cm}
\begin{subfigure}{.6\linewidth}
\hspace*{-1.8cm}
  \centering
  \includegraphics[scale=.32]{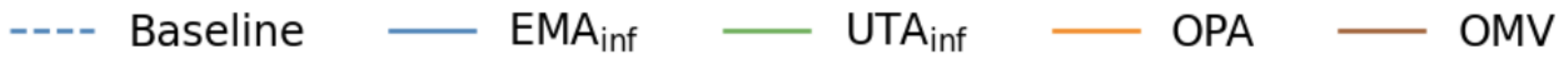}
\end{subfigure}
\vspace*{-.2cm}
\caption{Accuracy gains due to inference aggregations (Section~\ref{acc:inf_framework}) for DP-FTRL on original and PDS StackOverflow.}
\label{fig:stackoverflow}
\vspace*{-.1cm}
\end{figure*}

\subsection{Experiments with User-level Privacy on StackOverflow Dataset}\label{acc:user_exp}
In this section, we evaluate efficacy of our aggregation methods in a \emph{user-level DP} setting. Specifically, we first perform experiments with original StackOverflow data described in Section~\ref{sec:empEval:setup:dataset}, then describe a more real-world setting with periodically shifting distribution (PDS) of dataset and present results for the PDS setting.

\subsubsection{Aggregation Methods We Use With Original StackOverflow}\label{acc:user_exp:aggregation_methods}
We evaluate two \emph{training} and four \emph{inference} aggregation methods.
For training aggregations, we consider \algtr{EMA} and \algtr{UTA}  methods (Section~\ref{acc:tr_framework}).
For inference aggregations, we consider \alginf{EMA}, \alginf{UTA}, OPA, and OMV  methods  (Section~\ref{acc:inf_framework}).
Please refer to 
For \algtr{UTA}, we first use our adaptive training framework (ATF) with $f_{\sf UTA}$ as $f_{\sf AGG}$, as described in Section~\ref{acc:tr_framework}. Then we use our post-processing based inference framework on top of the checkpoints generated by ATF to produce the results in Tables~\ref{tab:stackoverflow} and~\ref{tab:cifar10} We similarly produce results for \algtr{EMA} in Tables~\ref{tab:stackoverflow} and~\ref{tab:cifar10}.
Following \citep{tanle19, de2022unlocking}, we use a warm-up schedule for the EMA coefficient as: $$\beta_t = \text{min}\left(\beta, ({1+t})/({10+t})\right)$$
Note that for EMA, one can further optimize this schedule and $\beta$, but note that widely increased tuning can have privacy consequences~\cite{papernot2021hyperparameter}. The other aggregations  have just one hyperparameter, $k$, making them more compute friendly.
All our results are average of 5 runs of each setting.

\subsubsection{Results for Original StackOverflow}\label{acc:user_exp:so_results}
In the rest of the paper, the tables present results for the final training round $T$,
while plots show results over the last $k$ rounds for some $k\ll T$. Due to large size of StackOverflow test data, we provide plots for accuracy on validation data and tables with accuracy on test data.

Table~\ref{tab:stackoverflow} presents the accuracy gains in  StackOverflow for $\epsilon\in\{\infty, 18.9, 8.2\}$ due to our training and inference aggregations. We observe that \textbf{our training aggregation \algtr{UTA} always provides the maximum accuracy gains}. Specifically, for $\epsilon$ of $\infty$, 18.9, and 8.2, \algtr{UTA} provides relative (absolute)  accuracy improvement over the baseline (DP-FTRL with momentum) of 2.97\% (0.75\%), 2.09\% (0.49\%), and 2.06\% (0.46\%) respectively. 
The corresponding relative (absolute)  accuracy improvement over $\emaBase$ (i.e., EMA over baseline with EMA coefficients as per \cite{de2022unlocking}) are 1.05\% (0.27\%), 1.48\% (0.45\%), and 1.43\% (0.32\%) respectively. Note that while \citet{de2022unlocking} do not have StackOverflow experiments, we provide results for $\emaBase$ using EMA and EMA coefficient $\beta$ suggested in~\citep{de2022unlocking}.

\begin{figure}[ht]
\centering
\hspace*{-.8cm}
\begin{subfigure}{\linewidth}
  \centering
  \includegraphics[scale=.35]{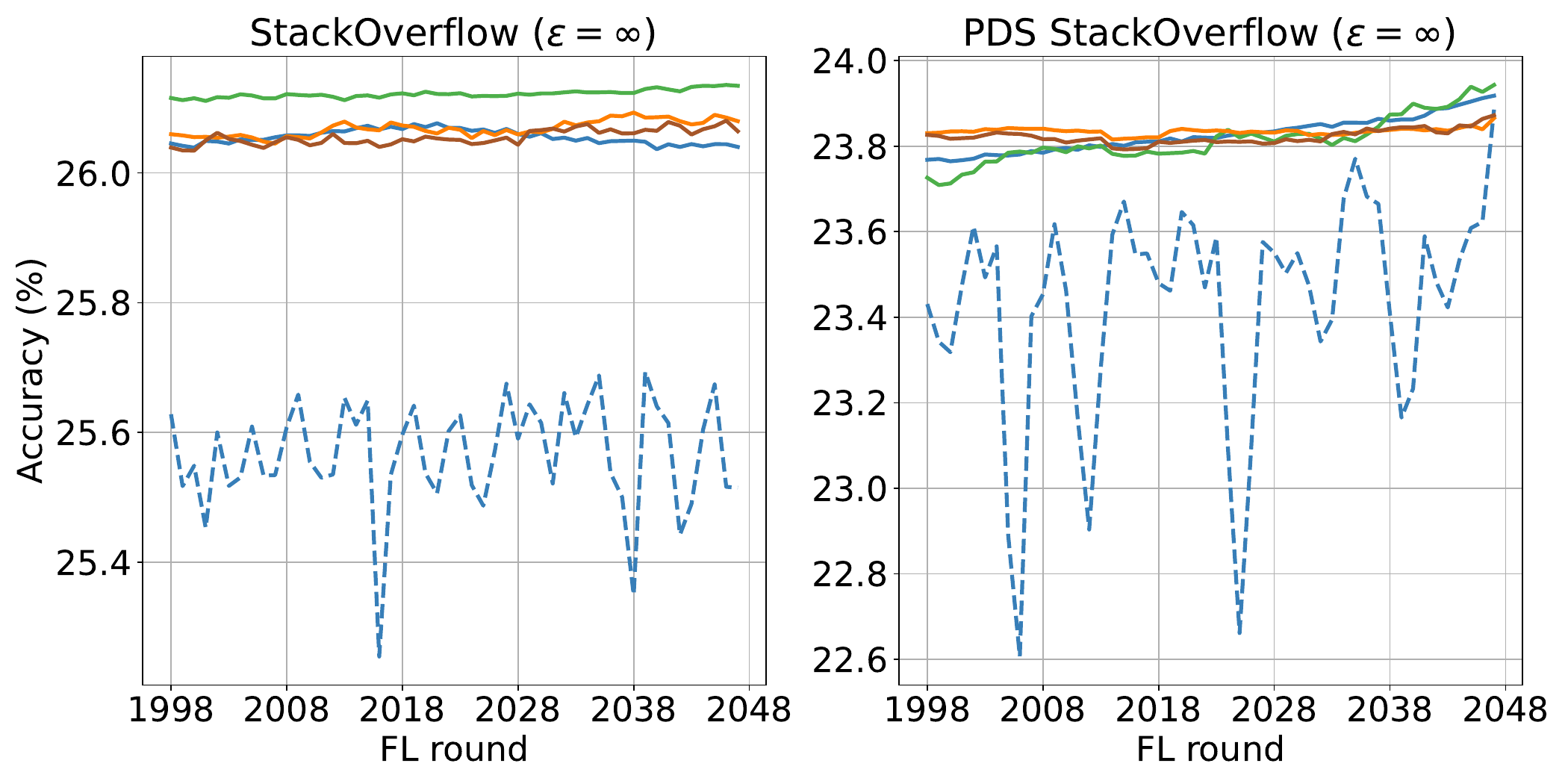}
\end{subfigure}%
\vspace*{-.1cm}
\begin{subfigure}{.6\linewidth}
\hspace*{-1.3cm}
  \centering
  \includegraphics[scale=.4]{figures/cb_legend.pdf}
\end{subfigure}
\vspace*{-.2cm}
\caption{Performances of inference aggregations (Section~\ref{acc:inf_framework}) in non-private settings ($\eps=\infty$). We note significant accuracy gains for DP-FTRL on original and PDS StackOverflow even in the non-private settings.}
\label{fig:so_non_private}
\end{figure}

Finally, in the leftmost two plots in Figure~\ref{fig:stackoverflow}, we focus on the inference aggregations since they just post-process the checkpoints of the state-of-the-art baseline run.  First, note that \textbf{all of inference aggregations significantly outperform the baseline (\alginf{UTA} performs the best among all inference aggregations)}.  Second, due to DP noise, the accuracy of baseline DP checkpoints has very high variance across training rounds, which is undesirable in practice.  However, we note that all considered inference aggregations significantly reduce such variance  while consistently providing gains in accuracy. In other words, \textbf{our checkpoints aggregations produce good DP models with high confidence, which is highly desired in practice}. 
The left plot in Figure~\ref{fig:so_non_private} presents results for the non-private setting with $\epsilon=\infty$ and we note similar improvements due to our inference aggregations.

It is worth mentioning that the DP state-of-the-art for the datasets we consider have repeatedly improved over the years since the foundational techniques from \cite{DP-DL} for CIFAR-10, and \cite{mcmahan2017learning} for StackOverflow, so we consider the consistent improvements that our proposed technique provide as significant improvements.

\begin{table*}[ht]
\caption{Test accuracy gains for original and periodic distribution shifting (PDS) CIFAR10. We present techniques from prior works (DP-SGD and $\emaBase$~\cite{de2022unlocking}) in .}
\label{tab:cifar10}
\centering
\vspace*{-.3cm}
\resizebox{\linewidth}{!}{%
\begin{tabular} {|c|c|c|c|c|c|c|c|c|}
    \hline
     {DP} &  & \multicolumn{2}{c|}{Training Aggregations} & \multicolumn{5}{c|}{Inference Aggregations} \\ \cline{3-9}
    $(\eps)$ &  & ${\sf EMA}_{\sf tr}$ & ${\sf UTA}_{\sf tr}$ &  & {${\sf EMA}_{\sf inf}$} & ${\sf UTA}_{\sf inf}$ & OPA & OMV \\ \hline
    \multicolumn{9}{|c|}{CIFAR10; DP-SGD; sample-level privacy} \\ \hline
    8 &  & 78.98 \tiny{$\pm$ 0.26} & \bf 79.96 \tiny{$\pm$ 0.24} &  & {79.41 \tiny{$\pm$ 0.51}} & 79.39 \tiny{$\pm$ 0.52} & 79.40 \tiny{$\pm$ 0.59} & 79.34 \tiny{$\pm$ 0.54} \\ \hline
    1 &  & 56.24 \tiny{$\pm$ 0.42} & \bf 57.51 \tiny{$\pm$ 0.31} &  & {56.61 \tiny{$\pm$ 0.91}} & 56.62 \tiny{$\pm$ 0.89} & 56.68 \tiny{$\pm$ 0.89} & 56.40 \tiny{$\pm$ 0.69}  \\ \hline
    \multicolumn{9}{|c|}{\emph{Periodic Distribution Shifting (PDS)} CIFAR10; DP-SGD; sample-level privacy} \\ \hline
    $8$ &  & 78.18 \tiny{$\pm$ 0.39} & \bf 79.19 \tiny{$\pm$ 0.44} &  & {78.24 \tiny{$\pm$ 0.92}} & 77.92 \tiny{$\pm$ 0.89} & 78.27 \tiny{$\pm$ 0.84} & 77.99 \tiny{$\pm$ 0.94} \\ \hline
    $1$ &  & 54.11 \tiny{$\pm$ 0.63} & \bf 55.01 \tiny{$\pm$ 0.48} &  & {54.04 \tiny{$\pm$ 0.81}} & 54.35 \tiny{$\pm$ 0.90} & 54.58 \tiny{$\pm$ 0.82} & 54.03 \tiny{$\pm$ 1.08} \\ \hline
\end{tabular}}
\vspace{-.2cm}
\end{table*}

\begin{figure*}[ht]
\centering
\begin{subfigure}{\linewidth}
  \centering
  \includegraphics[width=\linewidth, height=.25\linewidth]{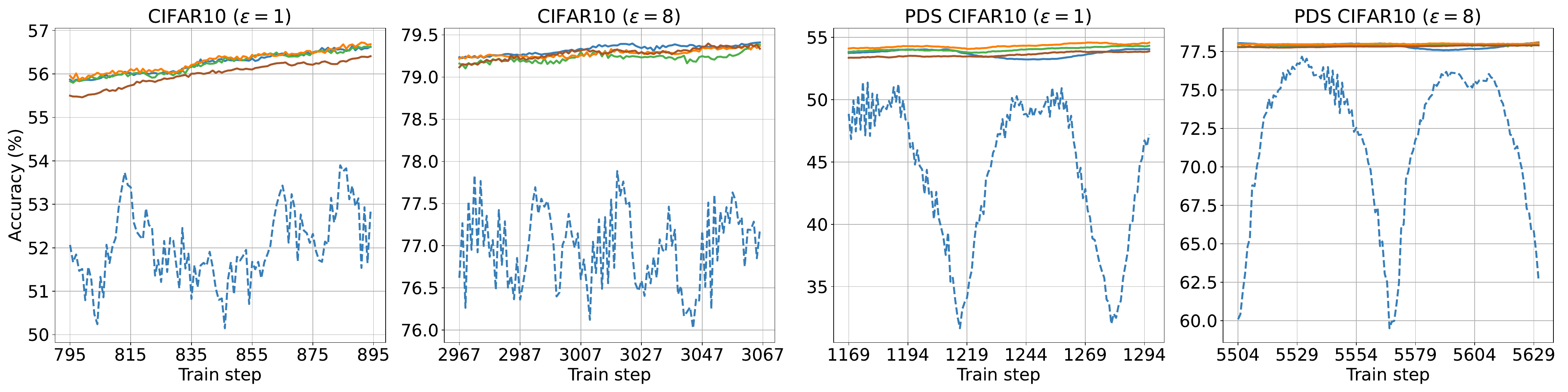}
\end{subfigure}%
\vspace*{-.01cm}
\begin{subfigure}{.6\linewidth}
  \centering
  \includegraphics[scale=.28]{figures/cb_legend.pdf}
\end{subfigure}
\vspace*{-.2cm}
\caption{Accuracy gains due to inference aggregation methods (Section~\ref{acc:inf_framework}) for DP-SGD on original and PDS CIFAR10.}
\label{fig:cifar10}
\end{figure*}

\subsubsection{Results for StackOverflow With Periodic Distribution Shifts}\label{acc:user_exp:pds_so_results}

Last four rows of Table~\ref{tab:stackoverflow} and the rightmost two plots of Figure~\ref{fig:stackoverflow} present accuracy gains for PDS StackOverflow (discussed in Section~\ref{acc:exp:pds}).
\textbf{For PDS StackOverflow as well, \algtr{UTA} always provides the maximum accuracy gains}; specifically for $\epsilon$ of $\infty$, 18.9, and 8.2, the relative (absolute) accuracy gains due to \algtr{UTA} over the DP-FTRL baseline are 1.55\% (0.37\%), 2.64\% (0.57\%), and 2.82\% (0.57\%)  respectively. While the relative (absolute) gains over $\emaBase$ are 1.67\% (0.42\%), 1.7\% (0.27\%), and 2.21\% (0.44\%) respectively. 
The rightmost two plots of Figure~\ref{fig:stackoverflow} show results of using our inference aggregations (Section~\ref{acc:inf_framework}) in PDS setting. We note that the variance of accuracy of the baseline DP-FTRL checkpoints is very high for the PDS setting, which is undesirable in practice. However, \textbf{our inference aggregations almost completely eliminate the variance in PDS setting, while producing more accurate predictions}.

\subsection{Experiments With Sample-level Privacy on CIFAR10 Dataset}\label{acc:sample_exp}

In this section, we evaluate efficacy of our aggregation methods (Section~\ref{acc:user_exp:aggregation_methods}) in a \emph{sample-level DP} setting with the original CIFAR10 and CIFAR10 with periodic distribution shifts (PDS). 

\subsubsection{Results for Original CIFAR10}\label{acc:sample_exp:cifar_results}
Table~\ref{tab:cifar10} and the left-most two plots in Figure~\ref{fig:cifar10} present the accuracy gains in CIFAR10 for $\epsilon\in\{1,8\}$. \textbf{For CIFAR10 as well \algtr{UTA} provides highest accuracy gains}.
Specifically, for $\eps$ of 1 and 8, the relative (absolute) accuracy gains due to \algtr{UTA} are 8.86\% (4.68\%) and 3.6\% (2.78\%) over the DP-SGD baseline, and they are 2.70\% (1.51\%) and 1.01\% (0.8\%) over $\emaBase$.
Among the inference aggregations, for $\epsilon=1$,  OPA provides the maximum relative (absolute) accuracy gain of 7.3\% (3.85\%), while for $\epsilon=8$, ${\sf EMA}_{\sf inf}$ provides maximum gain of 2.9\% (2.23\%) over the DP-SGD baseline.
We note from Figure~\ref{fig:cifar10} that \textbf{all checkpoints aggregations improve accuracy for all the training steps of DP-SGD for both $\epsilon$'s}.
Also note from Figure~\ref{fig:cifar10} that, the accuracy of baseline DP-SGD has a high variance across training steps and our inference aggregations significantly reduce this variance.

\begin{table*}[ht]
\caption{Test accuracy gains for original and periodic distribution shifting (PDS) CIFAR100. We present techniques from prior works (DP-SGD and $\emaBase$~\cite{de2022unlocking}) in .}
\label{tab:cifar100}
\centering
\vspace*{-.3cm}
\resizebox{\linewidth}{!}{%
\begin{tabular} {|c|c|c|c|c|c|c|c|c|}
    \hline
     {DP} &  & \multicolumn{2}{c|}{Training Aggregations} & \multicolumn{5}{c|}{Inference Aggregations} \\ \cline{3-9}
    $(\eps)$ &  & ${\sf EMA}_{\sf tr}$ & ${\sf UTA}_{\sf tr}$ &  & {${\sf EMA}_{\sf inf}$} & ${\sf UTA}_{\sf inf}$ & OPA & OMV \\ \hline
    \multicolumn{9}{|c|}{CIFAR100; DP-SGD; sample-level privacy} \\ \hline
    8 &  & 81.23 \tiny{$\pm$ 0.07} & \bf 81.54 \tiny{$\pm$ 0.08} &  & 80.88 \tiny{$\pm$ 0.10} &  80.83 \tiny{$\pm$ 0.09} & 80.92 \tiny{$\pm$ 0.10} &  80.82 \tiny{$\pm$ 0.10} \\ \hline
    1 &  & 75.58 \tiny{$\pm$ 0.09} & \bf 76.18 \tiny{$\pm$ 0.11} &  & {75.42 \tiny{$\pm$ 0.13}} &  75.62 \tiny{$\pm$ 0.12} & 75.51 \tiny{$\pm$ 0.16} & 75.57 \tiny{$\pm$ 0.18} \\ \hline
    \multicolumn{9}{|c|}{\emph{Periodic Distribution Shifting (PDS)} CIFAR100; DP-SGD; sample-level privacy} \\ \hline
    $8$ &  & 79.83 \tiny{$\pm$ 0.05} & \bf 81.27 \tiny{$\pm$ 0.06} &  & 80.53 {\tiny{$\pm$ 0.07}} &  80.53 \tiny{$\pm$ 0.08} & 80.49 \tiny{$\pm$ 0.08} &  80.41 \tiny{$\pm$ 0.09} \\ \hline
    $1$ &  &  74.88 \tiny{$\pm$ 0.09} & \bf 75.81 \tiny{$\pm$ 0.13} &  & {75.08 \tiny{$\pm$ 0.12}} & 75.81 \tiny{$\pm$ 0.16} & 75.01 \tiny{$\pm$ 0.17} &  74.97 \tiny{$\pm$ 0.18} \\ \hline
\end{tabular}}
\end{table*}

\newcommand{\pcvrres}[2]{#1, #2}
\begin{table*}[ht]
\caption{\emph{Relative} improvement in test \textit{AUC-loss} compared to DPSGD (No Agg) baseline for proprietary pCVR Dataset.
The two numbers presented for each algorithm are the improvements in the mean and standard deviation of the AUC-loss.
}
\label{tab:pCVR}
\centering
\vspace*{-.3cm}
\resizebox{\linewidth}{!}{%
\begin{tabular} {|c|c|c|c|c|c|c|c|}
\hline
{DP} &  \multicolumn{2}{c|}{Training Aggregations} & \multicolumn{5}{c|}{Inference Aggregations} \\ \cline{2-8}
$(\eps)$ &  ${\sf EMA}_{\sf tr}$ & ${\sf UTA}_{\sf tr}$ & & {${\sf EMA}_{\sf inf}$} & ${\sf UTA}_{\sf inf}$ & OPA & OMV \\ \hline
\multicolumn{8}{|c|}{pCVR; DP-SGD; sample-level privacy; (mean, std)} \\ \hline
6 &  \pcvrres{+0.32\%}{+18.9\%} & \textbf{\pcvrres{+0.53\%}{+26.2\%}} &  & \pcvrres{+0.22\%}{+7\%} & \pcvrres{+0.19\%}{+27.7\%} & \textbf{\pcvrres{+0.54\%}{+62.6\%}} & N/A  \\ \hline
\end{tabular}}
\end{table*}

\subsubsection{Results for CIFAR10 With Periodic Distribution Shifts}\label{acc:sample_exp:pds_cifar_results}
Section~\ref{acc:exp:pds} discusses how we emulate periodic distribution shifting (PDS) CIFAR10 data. Note that to train using DP-SGD on PDS CIFAR10, we set learning rate and noise multiplier, respectively, to 2 and 12 for $\epsilon=1$ and to 4 and 4 for $\epsilon=8$.

The last two rows of Table~\ref{tab:cifar10} show accuracy gains for PDS CIFAR10 due to our aggregation methods. 
As before, \textbf{the highest accuracy gains are due to our ${\sf UTA}_{\sf tr}$}.
Specifically, for $\eps$ of 1 and 8, the relative (absolute) accuracy gains due to \algtr{UTA} are 16.72\% (7.88\%) and 30.11\% (18.45\%) over the DP-SGD baseline, and they are, respectively, 1.79\% (0.97\%) and 1.53\% (1.2\%) over $\emaBase$.
Among the inference aggregations, OPA provides the maximum absolute accuracy gains over the DP-SGD baseline of 7.45\% and 17.37\%, respectively, for both $\eps\in\{1, 8\}$. From the rightmost two plots  (Figure~\ref{fig:cifar10}), we see that DP-SGD baseline models exhibit very large variance with PDS CIFAR10 across training steps, but all the inference aggregation methods completely eliminate the variance. 

Note that the improvements in PDS settings are significantly higher than that in the original settings, because the variance in model accuracy over training steps is large in PDS settings. Hence, the benefits of checkpoints aggregations magnify in these settings. For the PDS StackOverflow, where improvements are similar to StackOverflow,  we hypothesize that this might be due to the distributions in PDS CIFAR10 (completely different images from even/odd classes) being significantly farther apart compared to the distributions in PDS StacktOverflow (text from questions/answers).

\subsection{Experiments with Sample-level Privacy for CIFAR100 Dataset}
In this section, we evaluate our aggregation methods (Section~\ref{acc:user_exp:aggregation_methods}) in a \emph{sample-level DP} setting with the original CIFAR100 and CIFAR100 with periodic distribution shifts (PDS). 

\subsubsection{Improving CIFAR100 baseline}
First, we present a significant improvement over the SOTA baseline of~\cite{de2022unlocking}, i.e., ``No Agg" baseline in Table~\ref{tab:cifar100}). In particular, \emph{unlike in~\citep{de2022unlocking}, we fine-tune the final EMA checkpoint}, i.e., the one computed using EMA during pre-training over ImageNet. This results in major accuracy boosts of 5\% (70.3\% $\rightarrow$ 75.51\%) for $\epsilon=1$ and of 3.2\% (77.6\% $\rightarrow$ 80.81\%) for $\epsilon=8$ for the original CIFAR100 task. We obtain similarly high improvements by fine-tuning the EMA of pre-trained checkpoints (instead of the final checkpoint) for the PDS-CIFAR100 case. 
We emphasize that these gains are \emph{even before we use our aggregation methods}. We leave the further investigation of this phenomena to the future work. 

\subsubsection{Results for CIFAR100 and PDS CIFAR100}
We first discuss the gains for original CIFAR100 due to our aggregation methods; Table~\ref{tab:cifar100} shows the results. We note significant performance gains for CIFAR100 due to almost all of our aggregation methods. \textbf{For both $\epsilon\in\{1, 8\}$, \algtr{UTA} provides the highest accuracy gains}: For $\eps$ of 1 and 8, the relative (absolute) accuracy gains due to \algtr{UTA} are 0.89\% (0.67\%) and 0.91\% (0.73\%) over our improved DP-SGD baseline, and they are 1.4\% (1.05\%) and 0.82\% (0.66\%) over $\emaBase$.
Among the inference aggregations, for $\epsilon=1$,  \alginf{UTA} provides the maximum relative (absolute) accuracy gain of 0.15\% (0.11\%), while for $\epsilon=8$, OPA provides the gain of 0.14\% (0.11\%) over our improved DP-SGD baseline. 
The gains for CIFAR100 are seemingly smaller than those for CIFAR10, but as mentioned in Section~\ref{new_intro}, CIFAR100 with 100 classes is a much more difficult task, and hence, the accuracy gains in DP regime are notable.

For \textbf{PDS CIFAR100 task as well, \algtr{UTA} provides the highest accuracy gains}: For $\eps$ of 1 and 8, the relative (absolute) accuracy gains due to \algtr{UTA} are 7.0\% (4.97\%) and 5.33\% (4.11\%) over our improved DP-SGD baseline, and they are 1.87\% (1.4\%) and 0.92\% (0.74\%) over $\emaBase$.

\subsection{Experiments with Sample-level Privacy for pCVR}\label{acc:sample_exp_pcvr}

As this is a proprietary dataset, similar as prior works \citep{denison2022private,chua2024training}, 
we report only the \emph{relative} improvements in the AUC-loss; note that lower AUC-loss corresponds to better utility and improvement in AUC-loss means reduction in AUC-loss. The baseline we compare against is the model trained with DP-SGD (``No Agg"). The DP-SGD baseline has $< 5\%$ higher AUC-loss over the non-private model, which is similar to or slightly better than the DP-SGD models in prior work \cite{denison2022private,chua2024training}. Furthermore, as model stability is important for pCVR tasks, and DP training is well-known to increase variance, we also report the relative improvement in the standard deviation of the AUC-loss.

Table~\ref{tab:pCVR} presents the results. Similar to the other datasets, \textbf{all checkpoint aggregations improve AUC-loss}, i.e., reduce AUC-loss compared to the baseline. \algtr{EMA}, \algtr{UTA}, \alginf{UTA}, \alginf{OPA} also reduce the variance significantly.  
Among all aggregation methods, \alginf{OPA} provides the largest (relative) improvements in AUC-loss and its standard deviation of 0.54\% and 62.6\%, respectively, over the DP-SGD baseline. Notice that in the context of ads ranking, even 0.1\% relative improvement can have significant impact on revenue~\cite{wang2017deep}.

\section{Quantifying uncertainty due to differential privacy noise}\label{sec:uncertainty}

The prior literature on improving differentially private (DP) ML has focused on improving performances of DP models. However, a major issue with DP ML algorithms is high variance in their outputs due to high amounts of noise DP adds during training.
High variance in outputs, i.e., DP ML models, reduces the confidence of these models in their predictions which is undesired in practical applications. Hence, quantifying uncertainty in outputs of DP ML algorithms is instrumental towards success of DP ML in practice. 

Unfortunately, no prior work systematically investigates approaches for uncertainty quantification of DP deep learning. In this section, \emph{we propose the first method to quantify the uncertainty that the DP noise adds to the outputs of DP ML algorithms, without additional privacy cost or computation}.  In particular, we show that one can use the models along the path of DP-SGD to obtain an estimator for the variance introduced in the prediction due to the noise injected in the training process. 

For a bounded prediction function $f(\theta^{\sf DP-SGD})$ (with $\theta^{\sf DP-SGD}$ being the final model output by DP-SGD), a natural estimator of its variance is the ``independent runs estimator:'' running the algorithm independently $k$ times to obtain $\left\{f\left(\theta^{\sf DP-SGD}_1\right),\ldots,f\left(\theta^{\sf DP-SGD}_k\right)\right\}$, and then obtaining the sample variance of this set of predictions~\cite{brawner2018bootstrap}. However, the variance estimate is a post-processing of $k$ runs of DP-SGD, which means roughly speaking both its privacy and computational cost are $k$ times worse than DP-SGD. In particular, if we are restricted to one training of run of DP-SGD (e.g. due to computational costs), this method can only get one sample, i.e. the sample variance is undefined.

In this section, \emph{we demonstrate a variance estimator that can give an estimate using only a single run of DP-SGD, and also can outperform the independent runs estimator in some settings even when more than a single run is allowed.}

\subsection{Two Birds, One Stone: Our Uncertainty Estimator}

To address the two hurdles discussed above, we propose a simple yet efficient method that leverages intermediate checkpoints computed during a single run of DP-SGD. Specifically, we substitute the $k$ output models from the independent runs method with $k$ checkpoints from a single run. The rest of the confidence interval computation remains the same for both the methods.

We first give a theoretical upper bound on the error between the sample variance of a statistic calculated at $k$ intermediate checkpoints, and the true variance of this statistic at the final checkpoint. Our bias bound is decaying in two quantities: (i) the number of iterations $t_1$ before the first checkpoint, and (ii) $\gamma$, the minimum time between any two checkpoints.  
At a high level, our bound says that while checkpoints in DP-SGD are correlated, the addition of noise decreases their correlation over time, which justifies using them for uncertainty estimation in practice.  

Our bound, proved in \cref{sec:proofofcv}, is as follows:

\begin{thm}[Simplified version of Theorem~\ref{thm:checkpointvariance}]
Suppose $\calL(\theta;D)$ is  1-strongly convex and $M$-smooth, and $\sigma = 1$ in DP-SGD. Let $0<t_1<t_2<\ldots<t_k$ be such that $t_{i+1} \geq t_i + \gamma$ for $\forall i > 0$ and some minimum separation $\gamma$. Let $\{\theta_{t_i}: i \in [k]\}$ be the checkpoints, and $f: \Theta \rightarrow [-1, 1]$ be a statistic whose variance we wish to estimate. Let $V={\mathbf{Var}}\left[f(\theta_{t_k})\right]$, i.e. the variance of statistic at the final checkpoint (i.e., the final model), $\mu=\frac{1}{k}\sum\limits_{i=1}^k f(\theta_{t_i})$ be the sample mean, and $S=\left(\frac{1}{k-1}\sum\limits_{i=1}^k(f(\theta_{t_i})-\mu)^2\right)$ be the sample variance of the checkpoints. Then, for some ``burn-in'' times $\kappa_1, \kappa_2$ that are a function of $\theta_0, M, p$, we have:
$$|\mathbb{E}[S] - V| = \exp(-\Omega(\min\{t_1 - \kappa_1, \gamma - \kappa_2\})).$$
Here, the expectation $\mathbb{E}[\cdot]$ and the variance ${\mathbf{Var}}[\cdot]$ are over the randomness of DP-SGD.
\label{thm:checkpointvarianceR}
\end{thm}

\subsubsection{Proof Intuition}

To simplify the proof in \cref{sec:proofofcv} we actually prove a bound on the DP-LD algorithm, which is a continuous-time analog of DP-SGD. We defer a detailed discussion on the relationship between DP-LD and DP-SGD to \cref{sec:proofofcv}. For the following discussion, one should think of DP-LD and DP-SGD (with a small step size) as interchangeable.

Theorem~\ref{thm:checkpointvarianceR} and its proof say the following: (i) As we increase $t_1$, the time before the first checkpoint, each of the checkpoints' marginal distributions approaches the distribution of $\theta_{t_k}$, and (ii) As we increase $\gamma$, the time between checkpoints, the checkpoints' distributions approach pairwise independence. So increasing both $t_1$ and $\gamma$ causes our checkpoints to approach $k$ pairwise independent samples from the same distribution, i.e., our variance estimator approaches the true variance in expectation. 
To show both (i) and (ii), we build upon past results from the sampling literature to show a mixing bound of the following form: running DP-SGD from any point initialization $\theta_0$, the R\'enyi divergence between $\theta_t$ and the limit as $t \rightarrow \infty$ of DP-LD, $\theta_\infty$, decays exponentially in $t$. This mixing bound shows (i) since if $t_1$ is sufficiently large, then the distributions of all of $\theta_{t_1}, \theta_{t_2}, \ldots, \theta_{t_k}$ are close to $\theta_\infty$, and thus close to each other. This also shows (ii) since DP-LD is a Markov chain, i.e. the distribution of $\theta_{t_j}$ conditioned on $\theta_{t_i}$ is equivalent to the distribution of $\theta_{t_j - t_i}$ if we run DP-LD starting from $\theta_{t_i}$ instead of $\theta_0$. So our mixing bound shows that even after conditioning on $\theta_{t_i}$, $\theta_{t_j}$ has distribution close to $\theta_{\infty}$. Since $\theta_{t_j}$ is close to $\theta_\infty$ conditioned on any value of $\theta_{t_i}$, then $\theta_{t_j}$ is almost independent of $\theta_{t_i}$.

\mypar{Remark} In Theorem~\ref{thm:checkpointvarianceR}, $\kappa_1$ is a function of $\theta_0$ (the initialization model in DP-SGD) while $\kappa_2$ is independent of $\theta_0$. In particular, $\kappa_1$ can be arbitrarily large compared to $\kappa_2$ if $\theta_0$ is a poor choice for initialization, but we always have $\kappa_2 = O(\kappa_1)$. This implies the following:
\begin{itemize}
    \item When the initialization is poor, using the sample variance of the checkpoints as an estimator gives~\emph{a computational improvement} over the sample variance of $k$ independent runs of a training algorithm.
    
    \item Regardless of the initialization, using the sample variance of $k$ checkpoints is~\emph{never worse} in terms of computation cost than using $k$ independent runs.
    
    \item Checkpoints can provide tighter confidence intervals than independent runs under a fixed privacy constraint: 
    Suppose we have a fixed noise multiplier $\sigma / (L / n)$ we would like to use in training, as well as a fixed privacy budget. This implies we have a fixed number of iterations $T$ we can run. Fix $t_1$ and $\gamma$ such that the sample variance of the checkpoints has low bias; since $\kappa_1$ can be much larger than $\kappa_2$, we should also set $t_1$ to be much larger than $t_2$. Suppose we want to construct a confidence interval for a model trained for at least $t_1$ iterations. Using independent runs, we can get $T / t_1$ samples. Using  checkpoints from one $T$-iteration run, we can get $1 + \frac{T - t_1}{\gamma}$ samples. So we can get $\approx t_1 / \gamma$ times as many samples by using checkpoints, and thus get a narrower confidence interval under the same privacy budget.

\end{itemize}

\begin{figure}
\centering
\hspace*{-.3cm}
\includegraphics[scale=.4]{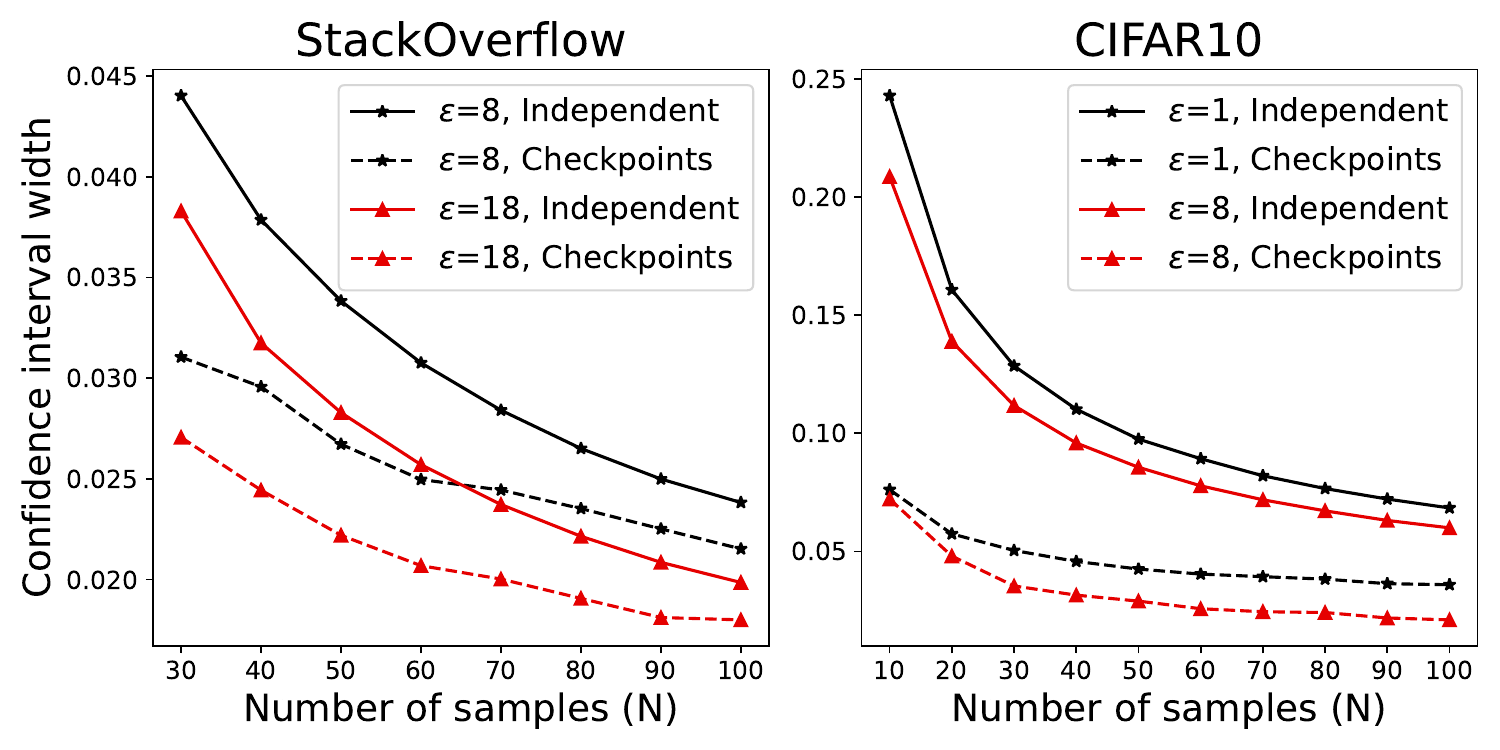}
\vspace*{-.3cm}
\caption{Uncertainty due to DP noise measured using confidence interval widths, computed via N bootstrap (independent) runs, and the last N checkpoints of a single run.}
\label{fig:uncertainty}
\end{figure}

\subsubsection{Empirical Analysis on Quadratic Losses}

We perform an empirical study of using the checkpoint variance estimator. We consider running DP-SGD on a 1-dimensional quadratic loss; we ignore clipping for simplicity, and assume the training rounds/privacy budget are fixed such that we can do exactly 128 rounds of DP-SGD. We set the learning rate $\eta = .07$, set the Gaussian variance such that the distribution of the final iterate has variance exactly 1, and set the initialization to be a random point drawn from $\calN(0, \sigma^2 = 100^2)$. Since $(1 - \eta)^{64} \approx 1/100$, under these parameters it takes roughly 64 rounds for DP-SGD to converge to within distance 1 of the minimizer. This reflects the setting where the burn-in time is a significant fraction of the training time, i.e. where \cref{thm:checkpointvarianceR} offers improvements over independent runs. We vary the burn-in time (i.e. round number of the first checkpoint) and the number of rounds between each checkpoint (i.e., the total number of checkpoints used) used in the variance estimator, and compute the error of the variance estimator across 1000 runs.

\begin{figure}
    \centering
    \includegraphics[width=0.5\linewidth]{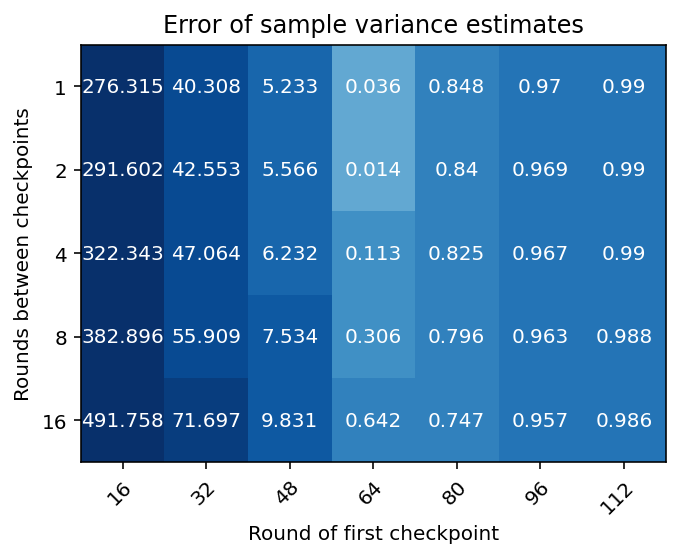}
    \vspace*{-.5cm}
    \caption{RMSE of the average sample variance given by the checkpoint estimator on quadratic losses.}
    \label{fig:quadratic}
\end{figure}

In Figure~\ref{fig:quadratic} we plot the RMSE of the variance estimator, which accounts for both the bias and variance of the estimator (note that \cref{thm:checkpointvarianceR} only looks at the bias; in \cref{sec:numcheckpoints} we discuss the problem of optimizing the checkpoints to minimize the RMSE). As predicted by \cref{thm:checkpointvarianceR}, we see that using too small a burn-in time causes a large bias, as the DP-SGD process has not had time to converge before the first checkpoint. We also see that using too large a burn-in time is suboptimal, since it reduces the number of checkpoints available to use in the estimator, increasing its variance. For rounds between checkpoints, at the best burn-in time of 64, we see it is best to choose 2 rounds between checkpoints. Again this matches the intuition of \cref{thm:checkpointvarianceR}: if we choose 1 round between checkpoints, checkpoints become too correlated which introduces bias into the variance estimate. At the same time, if we choose a larger separation like 16, we reduce the number of checkpoints the estimator uses, which increases the estimator's variance.

Recall that using independent runs of 128 iterations the independent runs' variance estimate is undefined, so all results in \cref{fig:quadratic} are improvements over that method. Even with e.g. 2 independent runs of 64 iterations, we only get 2 samples. Ignoring the bias due to using fewer iterations, the variance of this estimator is the variance of a degree-1 chi-squared distribution which is 2, i.e. it achieves RMSE at least 2.

\subsubsection{Empirical Analysis on Deep Learning}
We compare the uncertainty quantified using the independent runs method and using our method; experimental setup is the same as in Section~\ref{sec:empEval}. First, for a given dataset, we do 101 independent training runs (no budget split).
For accurately measuring the uncertainty of the training run at the specified privacy budget, we do not split the privacy budget across the independent runs here. Note that this is a superior baseline, as the overall privacy budget is significantly increased.
To compute uncertainty using the independent runs method for a fixed $N$, we first take the final model from $N$ of these runs (chosen randomly). 
Given an input sample, we compute prediction scores for each model, and compute the 95\% confidence interval width for the highest mean score.
We compute the average of the confidence interval widths in this manner for every sample from the validation set\footnote{Due to the large size of StackOverflow test data, we instead use validation data.}.
We conduct five independent repeats of this method, and report the mean confidence interval width as our final uncertainty estimate.
For computing uncertainty using our checkpoints based method, we do not optimize for the separation between checkpoints, giving a weaker hyperparameter-free method. we instead select the last $N$ checkpoints (i.e., last $N$ iterations) from a random training run, and obtain average confidence interval widths as above. T

Figure~\ref{fig:uncertainty} shows the results for StackOverflow and CIFAR10.
We see that the widths computed using  intermediate checkpoints consistently gives a reasonable lower bound on the widths computed using  independent runs, despite the strong baseline optimizing for the separation between checkpoints.
For instance, for DP-FTRL training on StackOverflow, the  confidence widths due to independent runs are always within a factor of 2 of the widths provided by our method across various privacy levels; for DP-SGD on CIFAR10, the bound is a factor is 4.

\section{Conclusions}
In this work, we design a general adaptive checkpoint aggregation framework to increase the performances of state-of-the-art DP ML techniques.  We show that uniform tail averaging of improves the excess empirical risk bound compared to the last checkpoint of DP-SGD. We demonstrate that uniform tail averaging \emph{during training} can provide significant improvements in prediction performances over the state-of-the-art for CIFAR10 and StackOverflow datasets, and the gains get magnified in more real-world settings with periodically varying training data distributions. Lastly, we prove that for some standard loss functions, the sample variance from last few checkpoints provides a good approximation of the variance of the final model of a DP run. Empirically, we show that the last few checkpoints can provide a reasonable lower bound for the variance of a converged DP model.

\bibliographystyle{plainnat}
\bibliography{reference}

\newpage
\appendix
\section{Details and Extensions for Theorem~\ref{thm:checkpointvarianceR}}

\subsection{Proof of Theorem \ref{thm:checkpointvarianceR}}\label{sec:proofofcv}

For completeness, we review the formal setup for the theorem we wish to prove. We focus on DP-LD, defined as follows:

\begin{equation}
    d\theta_t=-\nabla\calL(\theta_t;D)dt+\sigma \sqrt{2}dW_t.\label{eq:ld}
\end{equation}
One can view DP-LD and DP-SGD as approximations of each other as follows. We first reformulate (unconstrained) DP-SGD with step size $\eta$ as:
\[\tilde{\theta}_{(t+1)\eta} \leftarrow \tilde{\theta}_{t\eta} - \eta \nabla \calL(\tilde{\theta}_{t \eta}; D) + b_t, b_t \sim \calN(0, 2 \eta \sigma^2 \mathbb{I}_{p \times p}).\]

This reparameterization is commonly known as (DP-)SGLD~\cite{chourasia2021differential,ryffel2022differential,welling2011bayesian,zhang2017hitting}. Notice that we have reparameterized $\tilde{\theta}$ so that its subscript refers to the sum of all step-sizes so far, i.e. after $t$ iterations we have $\tilde{\theta}_{t\eta}$ and not $\tilde{\theta}_t$. Also notice that the variance of the noise we added is proportional to the step size $\eta$. In turn, for any $\eta$ that divides $t$, after $t/\eta$ iterations with step size $\eta$, the sum of variances of noises added is $2t\sigma^2$. This can be used to show a Renyi-DP guarantee for DP-SGLD with fixed $t$ that is independent of $\eta$, including in the limit as $\eta \rightarrow 0$.

Now, taking the limit as $\eta$ goes to $0$ of the sequence of random variables $\{\tilde{\theta}_{t \eta}\}_{t \in \mathbb{Z}_{\geq 0}}$ defined by DP-SGLD, we get a continuous sequence $\{\theta_t\}_{t \in \mathbb{R}_{\geq 0}}$. In particular, if we fix some $t$, then $\theta_t$ is the limit as $\eta$ goes to $0$ of $\tilde{\theta}_t$ defined by DP-SGLD with step size $\eta$. This sequence is exactly the sequence defined by DP-LD.

Note that the solutions $\theta_t$ to this equation are random variables. A key property of DP-LD is that the stationary distribution (equivalently, the limiting distribution as $t \rightarrow \infty$) has pdf proportional to $\exp(-\calL(\theta; D) / \sigma)$ under mild assumptions on $\calL(\theta;D)$ (which are satisfied by strongly convex and smooth functions).

While we focus on DP-LD for simplicity of presentation, a similar result can be proven for DP-SGLD. We discuss this in Section~\ref{sec:sgldextension}. 

To simplify proofs and presentation in the section, we will assume that (a) $\theta_0$ is a point distribution, (b) we are looking at unconstrained optimization over $\mathbb{R}^p$, i.e., there is no need for a projection operator in DP-SGD and DP-LD, (c) the loss $\calL$ is 1-strongly convex and $M$-smooth, and (d) $\sigma = 1$. We note that (a) can be replaced with $\theta_0$ being sampled from a random initialization without too much work, and (c) can be enforced for Lipschitz, smooth functions by adding a quadratic regularizer. We let $\theta^*$ refer to the (unique) minimizer of $\calL$ throughout the section.

Now, we consider the following setup: We obtain a single sample of the trajectory $\{\theta_t: t \in [0, T]\}$. We have some statistic $f:\Theta \rightarrow [-1, 1]$, and we wish to estimate the variance of some weighted average of the statistic across the checkpoints at times $0 < t_1 < t_2 < t_3 < \ldots < t_k = T$, i.e. the variance $V := \Var{\sum_i p_i f(\theta_{t_i})}$, where $\sum_i p_i = 1, p_i \geq 0$. To do so, we use a rescaling of the sample variance of the checkpoints. That is, our estimator is defined as $S = \frac{\sum_{i=1}^k p_i^2}{k-1} \sum_{i=1}^k (f(\theta_{t_i}) - \hmu)^2$ where $\hmu = \frac{1}{k} \sum_{i=1}^k f(\theta_{t_i})$.

\begin{thm}\label{thm:checkpointvariance}
Under the preceding assumptions/setup, for some sufficiently large constant $c$, let 
\[\kappa_1 = \frac{1}{2M} + \ln (c M (\ltwo{\theta_0 - \theta^*}^2 + p \ln(M))) + c \ln(1/\Delta),\]
\[\kappa_2 = \frac{1}{2M} + \ln (c M (\ln(1/\Delta) + p \ln(M))) + c \ln(1/\Delta),\] (recall that $p$ is the dimensionality of the space). Then, if $t_1 > \kappa_1$ and $t_{i+1} > t_i + \kappa_2$ for all $i > 0$, for $S, V$ as defined above:

\[|\mathbb{E}[S] - V| = O(\Delta \sum_{i=1}^k p_i^2).\]
\end{thm}

Theorem~\ref{thm:checkpointvariancesimple} is the special case of setting $p_k = 1$ and $p_i = 0, i \neq k$. Note that $\kappa_1$ can be arbitrarily large compared to $\kappa_2$ due to its dependence on $\theta_0$, whereas $\kappa_2 = O(\kappa_1)$. In particular, $\kappa_1 + (k-1)\kappa_2$ (the time to do one long run and use $k$ intermediate checkpoints for uncertainty estimation) can be significantly smaller than $k\kappa_1$ (the time to do $k$ independent runs and use the final checkpoints for uncertainty estimation). Before proving this theorem, we need a few helper lemmas about R\'enyi divergences:

\begin{definition}
The R\'enyi divergence of order $\alpha > 1$ between two distributions $\calP$ and $\calQ$ (with support $\mathbb{R}^d$), $D_\alpha(\calP || \calQ)$, is defined as follows:
\[D_{\alpha}(\calP || \calQ) := \int_{\theta \in \mathbb{R}^d} \frac{P(\theta)^\alpha}{Q(\theta)^{\alpha-1}}d\theta\]
\end{definition}

We refer the reader to e.g.~\cite{vanErvenRenyi, mironov2017renyi} for properties of the R\'enyi divergence. The following property shows that for any two random variables close in R\'enyi divergence, functions of them are close in expectation:

\begin{lem}\label{lem:differenceofexpectations}[Adapted from Lemma C.2 of~\cite{bun2016concentrated}]
Let $\calP$ and $\calQ$ be two distributions on $\Omega$ and $g:\Omega\to[-1, 1]$. Then,
$$\left|\mathbb{E}_{x\sim\calP}\left[g(x)\right]-\mathbb{E}_{x\sim\calQ}\left[g(x)\right]\right|\leq\sqrt{e^{D_2(\calP||\calQ)}-1}.$$
Here, $D_2(\calP||\calQ)$ corresponds to R\'enyi divergence of order two between the distributions $\calP$ and $\calQ$.
\label{lem:struct1}
\end{lem}

The next lemma shows that the solution to DP-LD approaches $\theta_\infty$ exponentially quickly in R\'enyi divergence.

\begin{lem}\label{lem:renyifrompoint}
Fix some point $\theta_0$.  Assume $\calL$ is 1-strongly convex, and $M$-smooth. Let $\calP$ be the distribution of $\theta_{t}$ according to DP-LD for $\sigma = 1$ and:

\[t := 1/2M + \ln(c(M \ltwo{\theta_0-\theta^*}^2 + p \ln (M))) + c \ln(1/\Delta).\]

Where $c$ is a sufficiently large constant. Let $Q$ be the stationary distribution of DP-LD. Then:
\[D_2(\calP || \calQ) = O (\Delta^2).\]
\end{lem}

The proof of this lemma builds upon techniques in \cite{ganesh2020langevin}, and we defer it to the appendix. Our final helper lemma shows that $\theta_\infty$ is close to $\theta^*$ with high probability:

\begin{lem}\label{lem:lsctailbound}
Let $\theta_\infty$ be the random variable given by the stationary distribution of DP-LD for $\sigma = 1$. If $\calL$ is 1-strongly convex, then:

\[\Pr[\ltwo{\theta_\infty - \theta^*} > \sqrt{p} + x ] \leq \exp(-x^2/2).\]
\end{lem}
\begin{proof}
We know the stationary distribution has pdf proportional to $\exp(-\calL(\theta_t; D))$. In particular, since $\calL$ is 1-strongly convex, this means $\theta_\infty$ is a sub-Gaussian random vector (i.e., its dot product with any unit vector is a sub-Gaussian random variable), and thus the above tail bound applies to it.
\end{proof}

We now will show that under the assumptions in Theorem~\ref{thm:checkpointvariance}, every checkpoint is close to the stationary distribution, and that every pair of checkpoints is nearly pairwise independent.

\begin{lem}\label{lem:nearlyindependent}
Under the assumptions/setup of Theorem~\ref{thm:checkpointvariance}, we have:

\begin{enumerate}
    \item[(E1)] $\forall i: |\mathbb{E}[(f(\theta_{t_i}))] - \mathbb{E}[(f(\theta_{t_k}))]| = O(\Delta)$,
    \item[(E2)]$\forall i: |\mathbb{E}[(f(\theta_{t_i})^2)] - \mathbb{E}[f(\theta_{t_k})^2]| = O(\Delta)$,
    \item[(E3)] $\forall i < j: |\Cov{f(\theta_{t_i})}{f(\theta_{t_j})}| = O(\Delta)$.
\end{enumerate}
\end{lem}
\begin{proof} 
We assume without loss of generality $\Delta$ is at most a sufficiently small constant; otherwise, since $f$ has range $[-1, 1]$, all of the above quantities can easily be bounded by 2, so a bound of $O(\Delta)$ holds for any distributions on $\{\theta_{t_i}\}$.

For (E1), by triangle inequality, it suffices to prove a bound of $O(\Delta)$ on $|\mathbb{E}[f(\theta_{t_i})] - \mathbb{E}[f(\theta_\infty)]|$. We abuse notation by letting $\theta_t$ denote both the random variable and its distribution. Then:

\[|\mathbb{E}[f(\theta_{t_i})] - \mathbb{E}[f(\theta_\infty)]| \]\[\stackrel{\text{Lemma~\ref{lem:differenceofexpectations}}}{\leq} \sqrt{e^{D_2(f(\theta_{t_i}), f(\theta_\infty))} - 1} \stackrel{(\ast_1)}{\leq}\sqrt{e^{D_2(\theta_{t_i}, \theta_\infty)} - 1}\]\[ \stackrel{\text{Lemma~\ref{lem:renyifrompoint}}, t_i \geq \kappa_1}{=} \sqrt{e^{O(\Delta^2)} - 1} \stackrel{(\ast_2)}{=} O(\Delta).\]

In $(\ast_1)$ we use the data-processing inequality (Theorem 9 of \cite{vanErvenRenyi}), and in $(\ast_2)$ we use the fact $e^{x} - 1 \leq 2x, x \in [0, 1]$ and our assumption on $\Delta$.

(E2) follows from (E1) by just using $f^2$ (which is still bounded in $[-1, 1]$) instead of $f$.

For (E3), note that since DP-LD is a (continuous) Markov chain, the distribution of $\theta_{t_j}$ conditioned on $\theta_{t_i}$ is the same as the distribution of $\theta_{t_j - t_i}$ according to DP-LD if we start from $\theta_{t_i}$ instead of $\theta_0$. Let $\calP$ be the joint distribution of $\theta_{t_i}, \theta_{t_j}$. Let $\calQ$ be the joint distribution of $\theta_{t_i}, \theta_\infty$ (since DP-LD has the same stationary distribution regardless of its initialization, this is a pair of independent variables). Let $\calP', \calQ'$ be defined identically to $\calP || \calQ$, except when sampling $\theta_{t_i}$, if $\ltwo{\theta_{t_i} - \theta^*} > \sqrt{p} + \sqrt{2 \ln(1/\Delta)}$ we instead set $\theta_{t_i} = \theta^*$ (and in the case of $\calP'$, we instead sample $\theta_{t_j}$ from $\theta_{t_j} | \theta_{t_i} = \theta^*$ when this happens). Let $\calR$ denote this distribution over $\theta_{t_i}$. Then similarly to the proof of (E1) we have:
\begin{align*}
|\mathbb{E}_{\calP'}&[f(\theta_{t_i})f(\theta_{t_j})] - \mathbb{E}_{\calQ'}[f(\theta_{t_i})]\mathbb{E}[f(\theta_{\infty})]|\\ &\stackrel{\text{Lemma~\ref{lem:differenceofexpectations}}}{\leq} \sqrt{e^{D_2(\calP', \calQ')} - 1} \\
&\stackrel{(\ast_3)}{\leq}  \sqrt{e^{\max_{\theta_{t_i} \in \text{supp}(\calR)}\{D_2(\theta_{t_j} | \theta_{t_i}, \theta_\infty)\}} - 1}.\\
&\stackrel{\text{Lemma~\ref{lem:renyifrompoint}}, t_j-t_i \geq \kappa_2}{=} \sqrt{e^{O(\Delta^2)} - 1} = O(\Delta).
\end{align*}

Here $(\ast_3)$ follows from the convexity of R\'enyi divergence, and in our application of \ref{lem:renyifrompoint}, we are using the fact that for all $\theta_{t_i} \in \text{supp}(\calR)$, $\ltwo{\theta_{t_i} - \theta^*} \leq \sqrt{p} + \sqrt{2 \ln(1/\Delta)}$. Furthermore, by Lemma~\ref{lem:lsctailbound}, we know $\calP$ and $\calP'$ (resp. $\calQ$ and $\calQ'$) differ by at most $\Delta$ in total variation distance. So, since $f$ is bounded in $[-1, 1]$, we have:

\[|\mathbb{E}_{\calP}[f(\theta_{t_i})f(\theta_{t_j})] - \mathbb{E}_{\calP'}[f(\theta_{t_i})f(\theta_{t_j})]| \leq \Delta,\]
\[|\mathbb{E}_{\calQ}[f(\theta_{t_i})]\mathbb{E}[f(\theta_\infty)] - \mathbb{E}_{\calQ'}[f(\theta_{t_i})]\mathbb{E}[f(\theta_\infty)]| \leq \Delta.\]

Then by applying triangle inequality twice:

\[|\mathbb{E}_{\calP}[f(\theta_{t_i})f(\theta_{t_j})] - \mathbb{E}_{\calQ}[f(\theta_{t_i})]\mathbb{E}[f(\theta_{\infty})]| = O(\Delta)\]

Now we can prove (E3) as follows:

\begin{align*}
&|\Cov{f(\theta_{t_i})}{f(\theta_{t_j})}|\\
&=|\mathbb{E}[(f(\theta_{t_i})-\mathbb{E}[f(\theta_{t_i})])(f(\theta_{t_j})-\mathbb{E}[f(\theta_{t_j})])]|\\
&=|\mathbb{E}[f(\theta_{t_i})f(\theta_{t_j})]-\mathbb{E}[f(\theta_{t_i})]\mathbb{E}[f(\theta_{t_j})]|\\
&\leq|\mathbb{E}[f(\theta_{t_i})f(\theta_{t_j})]-\mathbb{E}[f(\theta_{t_i})]\mathbb{E}[f(\theta_{\infty})]| + \\ & \qquad \ \ |\mathbb{E}[f(\theta_{t_i})]\mathbb{E}[f(\theta_{\infty})] - \mathbb{E}[f(\theta_{t_i})]\mathbb{E}[f(\theta_{t_j})]|\\
&\leq O(\Delta) + |\mathbb{E}[f(\theta_{\infty})] - \mathbb{E}[f(\theta_{t_j})]| = O(\Delta).
\end{align*}

\end{proof}

\begin{proof}[Proof of Theorem~\ref{thm:checkpointvariance}]
We again assume without loss of generality $\Delta$ is at most a sufficiently small constant. 
The proof strategy will be to express $\mathbb{E}[S]$ in terms of individual variances $\Var{f(\theta_{t_i})}$, which can be bounded using Lemma~\ref{lem:nearlyindependent}.

We have the following:
\begin{align}
    &\mathbb{E}[S]=\frac{\sum_{i=1}^k p_i^2}{k-1}\sum\limits_{i=1}^k\mathbb{E}\left[(f(\theta_{t_i})-\hmu)^2\right] \nonumber\\
    &=\frac{\sum_{i=1}^k p_i^2}{k-1}\sum\limits_{i=1}^k\mathbb{E}\left[\left(\frac{k-1}{k}\right)^2\left(\underbrace{f(\theta_{t_i})}_{x_i}-\underbrace{\frac{1}{k-1}\sum\limits_{j\in[k],j\neq i}f(\theta_{t_j})}_{y_i}\right)^2\right].
    \label{eq:3}
\end{align}
From~\eqref{eq:3}, we have the following:
\begin{align}
    &\mathbb{E}\left[(x_i-y_i)^2\right]\nonumber\\
    &=\mathbb{E}[x_i^2]-2\mathbb{E}[x_iy_i]+\mathbb{E}[y_i^2]\nonumber\\
    &=\left(\mathbb{E}[x_i^2]-\left(\mathbb{E}[x_i]\right)^2\right)+\left(\mathbb{E}[y_i^2]-\left(\mathbb{E}[y_i]\right)^2\right)+\nonumber\\
    &\qquad\qquad\left(\left(\mathbb{E}[x_i]\right)^2+\left(\mathbb{E}[y_i]\right)^2-2\mathbb{E}\left[x_iy_i\right]\right)\nonumber\\
    &=\underbrace{\Var{x_i}}_A+\underbrace{\Var{y_i}}_{B}+\underbrace{\left(\left(\mathbb{E}[x_i]\right)^2+\left(\mathbb{E}[y_i]\right)^2-2\mathbb{E}\left[x_iy_i\right]\right)}_C.
    \label{eq:4}
\end{align}
In the following, we bound each of the terms $A$, $B$, and $C$ individually. First, let us consider the term $B$. We have the following:
    
\begin{align}
    &B=\Var{y_i}=\frac{1}{(k-1)^2}\nonumber\\
    &\left(\sum\limits_{j\in[k],j\neq i}\Var{f(\theta_{t_j})}+2\sum\limits_{\substack{1\leq j< \ell\leq k \\ j\neq i,\ell\neq i}}\Cov{f(\theta_{t_j})}{f(\theta_{t_\ell})}\right).
    \label{eq:5}
\end{align}

Plugging Lemma~\ref{lem:nearlyindependent}, (E3) into~\eqref{eq:5} we bound the variance of $y_i$ as follows:
\begin{equation}
    B=\Var{y_i}=\frac{1}{(k-1)^2}\left(\sum\limits_{j\in[k],j\neq i}\Var{f(\theta_{t_j})}\right)\pm O(\Delta).\label{eq:12}
\end{equation}
We now focus on bounding the term $C$ in~\eqref{eq:4}. Lemma~\ref{lem:nearlyindependent}, (E1) and (E3) implies the following:
\begin{align}
    (\mathbb{E}[x_i])^2&= (\mathbb{E}[f(\theta_{t_k})])^2 \pm O(\Delta),\label{eq:13}\\
    (\mathbb{E}[y_i])^2&= (\mathbb{E}[f(\theta_{t_k})])^2 \pm O(\Delta),\label{eq:14}\\
    \mathbb{E}[x_i y_i]&= (\mathbb{E}[f(\theta_{t_k})])^2+O(\Delta).\label{eq:15}
\end{align}
Plugging~\eqref{eq:13},\eqref{eq:14}, and~\eqref{eq:15} into~\eqref{eq:4}, we have
\begin{align}
    &\mathbb{E}\left[(x_i-y_i)^2\right]\nonumber\\ 
    &= \Var{f(\theta_{t_i})}+\frac{1}{(k-1)^2}\left(\sum\limits_{j\in[k],j\neq i}\Var{f(\theta_{t_j})}\right)\pm O(\Delta)\label{eq:16}.
\end{align}
Now, Lemma~\ref{lem:nearlyindependent}, (E1) and (E2) implies $$\forall i,: \left|\Var{f(\theta_{t_i})} - \frac{V}{\sum_{i=1}^k p_i^2}\right| = O(\Delta)$$. So from \eqref{eq:16} we have the following: 
\begin{equation}
    \mathbb{E}\left[(x_i-y_i)^2\right]= V \cdot \frac{k}{(k-1)\sum_{i=1}^k p_i^2}\pm O(\Delta)\label{eq:17}.
\end{equation}
Plugging this bound back in~\eqref{eq:3}, we have the following:
\begin{align}
    \mathbb{E}[S] &= \frac{\sum_{i=1}^k p_i^2}{k-1}\cdot\left(\frac{k-1}{k}\right)^2\cdot k \cdot  \left(V \cdot \frac{k}{(k-1) \sum_{i=1}^k p_i^2} \pm O(\Delta)\right)\nonumber\\
    &=V \pm O(\Delta \sum_{i=1}^k p_i^2).\label{eq:18}
\end{align}
Which completes the proof.
\end{proof}

\subsection{Optimizing the Number of Checkpoints}\label{sec:numcheckpoints}

In Theorem~\ref{thm:checkpointvariance}, we fixed the number of checkpoints and gave lower bounds on the burn-in time and separation between checkpoints needed for the sample variance bound to have bias at most $\Delta$. We could instead consider the problem where $T$, the time of the final checkpoint, is fixed, and we want to choose $k$ which minimizes the (upper bound on) mean squared error of the sample variance of $\{f(\theta_{iT/k})\}_{i \in [k]}$. Here, we sketch a solution to this problem using the bound from this section.

The mean squared error of the sample variance is the sum of the bias and variance of this estimator. We will use the following simplified reparameterization of Theorem~\ref{thm:checkpointvariance}:

\begin{thm}[Simpler version of Theorem~\ref{thm:checkpointvariance}]\label{thm:checkpointvariancesimple}
Let $c_1 := \frac{1}{2M} + \ln (c_2 M (p + \ltwosq{\theta_0 - \theta^*}))$, where $c_2$ is a sufficiently large constant. Then if $S$ is the sample variance of $\{f(\theta_{iT/k})\}_{i \in [k]}$, $V$ is the true variance of $f(\theta_T)$, and $T/k > c_1$:

\[|\mathbb{E}[S] - V|^2 \leq \exp\left(-\frac{T/k - c_1}{c_2}\right).\]
\end{thm}

One can also bound the variance of $S$:

\begin{lem}
If $\bar{S}$ is the sample variance of $k > 1$ i.i.d. samples of $\theta_T$, then if $c_2$ is a sufficiently large constant, for $c_1$ as defined in \cref{thm:checkpointvariancesimple}:

\[\Var{\bar{S}} \leq \frac{1}{k}, |\Var{S} - \Var{\bar{S}}| \leq 2 \exp\left(-\frac{T/k - c_1}{c_2}\right).\]

\end{lem}
\begin{proof}
Let $x_1, \ldots, x_k$ be $k$ i.i.d. samples of $f(\theta_T)$, then since each $x_i$ is in the interval $[-1, 1]$:

\[\Var{\bar{S}} = \frac{\mathbb{E}[x_1^4]}{k} - \frac{\Var{x_1}(k-3)}{k(k-1)} \leq \frac{1}{k}.\]

Giving the first part of the lemma. For the second part, let $x_i$ be the sampled value of $f(\theta_{iT/k})$. Then:

\[\mathbb{E}[S^2] = \mathbb{E}\left[\left(\frac{1}{k-1}\sum_{i \in [k]} \left(x_i - \frac{1}{k} \sum_{j \in [k]} x_j \right)^2\right)^2\right].\] For some coefficients $c_{i,j,\ell,m}$, this can be written as $$\sum_{i \leq j \leq \ell \leq m} c_{i,j,\ell,m} \mathbb{E}[x_i x_j x_\ell x_m]$$ where $\sum_{i \leq j \leq \ell \leq m} |c_{i,j,\ell,m}| \leq 2$. By a similar argument to Theorem~\ref{thm:checkpointvariance}, the change in this expectation if we instead use $x_i$ that are i.i.d. is then at most $\exp\left(-\frac{T/k - c_1}{c_2}\right)$ as long as $c_2$ is a sufficiently large constant. In other words, $|\mathbb{E}[S^2] - \mathbb{E}[\bar{S}^2]| \leq \exp\left(-\frac{T/k - c_1}{c_2}\right)$. A similar argument applies to $E[S]^2$, giving the second part of the lemma.
\end{proof}

Putting it all together, we have an upper bound on the mean squared error of the sample variance of: 

\[\frac{1}{k} + 3 \exp\left(-\frac{T/k - c_1}{c_2}\right),\]

Assuming $k > 1, T/k > c_1$. Minimizing this expression with respect to $k$ gives
\[k = \frac{T}{c_1  + c_2 \ln (3T / c_2)},\]

which we can then round to the nearest integer larger than 1 to determine the number of checkpoints to use that minimizes our upper bound on the mean squared error. Of course, if $T < 2c_1$ then Theorem~\ref{thm:checkpointvariance} cannot be applied to give a meaningful bias bound for any number of checkpoints, so this choice of $k$ is not meaningful in that case.

\subsection{Proof of Lemma~\ref{lem:renyifrompoint}}
We will bound the divergences $D_\alpha(P_1|| P_2), D_\alpha(P_2|| P_3), D_\alpha(P_3|| P_4)$ where $P_1$ is the distribution $\theta_\eta$ that is the solution to \eqref{eq:ld}, $P_2$ is a Gaussian centered at the point $\theta_0 - \eta \nabla \calL(\theta_0; D)$, $P_3$ is a Gaussian centered at $\theta^*$, and $P_4$ is the stationary distribution of \eqref{eq:ld}. Then, we can use the approximate triangle inequality for R\'enyi divergences to convert these pairwise bounds into the desired bound.

\begin{lem}\label{lem:p1p2divergence}
Fix some $\theta_0$. Let $P_1$ be the distribution of $\theta_\eta$ that is the solution to \eqref{eq:ld}, and let $P_2$ be the distribution $N(\theta_0 -  \eta \nabla \calL(\theta_0; D), 2\eta)$. Then:

\[D_\alpha(P_1|| P_2) = O\left(M^2 \ln(\alpha) \cdot \max\{p \eta^2, \ltwo{\theta_0-\theta^*}^2 \eta^3\}\right)\]
\end{lem}
\begin{proof}
Let $\theta_t$ be the solution trajectory of \eqref{eq:ld} starting from $\theta_0$, and let $\theta_t'$ be the solution trajectory if we replace $\nabla \calL(\theta_t; D)$ with $\nabla \calL(\theta_0; D)$. Then $\theta_\eta$ is distributed according to $P_1$ and $\theta_\eta'$ is distributed according to $P_2$.

By a tail bound on Brownian motion (see e.g. Fact 32 in \cite{ganesh2020langevin}), we have that $\max_{t \in [0, \eta]} \ltwo{\int_0^t dW_s ds} \leq \sqrt{\eta (p + 2 \ln(2/\delta))}$ w.p. $1-\delta$. Then following the proof of Lemma 13 in \cite{ganesh2020langevin}, w.p. $1-\delta$, \[\max_{t \in [0, \eta]} \ltwo{\theta_t - \theta_0} \leq cM(\sqrt{p}+\sqrt{\ln(1/\delta)})\sqrt{\eta} + M \ltwo{\theta_0 - \theta^*}\eta ,\] for some sufficiently large constant $c$, and the same is true w.p. $1-\delta$ over $\theta_t'$. Now, following the proof of Theorem 15 in \cite{ganesh2020langevin}, for some constant $c'$, we have the divergence bound $D_\alpha(P_1|| P_2) \leq \epsilon$ as long as:

\[\frac{M^4 \ln^2 \alpha}{\epsilon^2} (p\eta^2 + \ltwosq{\theta_0 - \theta^*} \eta^3) < c'.\]

In other words, for any fixed $\eta$, we get a divergence bound of:

\[D_\alpha(P_1|| P_2) = O\left(M^2 \ln(\alpha) \cdot \max\{p \eta^2, \ltwo{\theta_0-\theta^*}^2 \eta^3\}\right),\]

as desired.
\end{proof}

\begin{lem}\label{lem:p2p3divergence}
Let $P_2$ be the distribution $N(\theta_0 -  \eta \nabla \calL(\theta_0; D), 2\eta)$ and $P_3$ be the distribution $N(\theta^*, 2\eta)$. Then for $\eta \leq 2/ M$:

\[D_\alpha(P_2|| P_3) \leq \frac{\alpha \ltwosq{\theta_0 - \theta^*}}{4\eta}.\]
\end{lem}
\begin{proof}
By contractivity of gradient descent we have:
\[\ltwo{\theta_0 - \eta \nabla \calL(\theta_0; D) - \theta^*} \leq \ltwo{\theta - \theta^*}.\]

Now the lemma follows from R\'enyi divergence bounds between Gaussians (see e.g., Example 3 of \cite{vanErvenRenyi}).
\end{proof}

\begin{lem}\label{lem:p3p4divergence}
Let $P_3$ be the distribution $N(\theta^*, 2\eta)$ and let $P_4$ be the stationary distribution of \eqref{eq:ld}. Then for $\eta \leq 1/2M$ we have:
\[D_\alpha(P_3|| P_4) \leq \frac{\alpha}{\alpha-1}\left(\frac{p}{2} \ln(1/\eta) - \ln(2\pi)\right) + \frac{p}{2} \ln (\alpha/4\pi\eta).\]
\end{lem}
\begin{proof}

We have $P_3(\theta) = P_3(\theta^*) \exp(-\frac{1}{4\eta} \ltwosq{\theta - \theta^*})$ where $P_3(\theta^*) = \left(\frac{1}{4\pi \eta}\right)^d$. By $M$-smoothness of the negative log density of $P_4$, we also have $P_4(\theta) \geq P_4(\theta^*) \exp(-\frac{M}{2} \ltwosq{\theta - \theta^*})$. In addition, since $P_4$ is $1$-strongly log concave, $P_4(\theta^*) \geq \left(\frac{1}{2\pi}\right)^{p/2}$ (as the $1$-strongly log concave density with mode $\theta^*$ that minimizes $P_4(\theta^*)$ is the multivariate normal with mean $\theta^*$ and identity covariance). Finally, for $\alpha \geq 1$ and $\eta \leq 1/2M$, we have $\alpha / 4 \eta > (\alpha - 1)M/2$. Putting it all together:

\begin{align}
&\exp((\alpha-1) D_\alpha(P_3|| P_4)) \\
&= \int \frac{P_3(\theta)^\alpha}{P_4(\theta)^{\alpha-1}} d\theta \\
&= \frac{P_3(\theta^*)^\alpha}{P_4(\theta^*)^{\alpha-1}} \int \exp\left(- (\frac{\alpha}{4\eta} - (\alpha - 1)\frac{M}{2}) \ltwosq{\theta - \theta^*}\right) d\theta \\
\leq \left(\frac{1}{4\pi\eta}\right)^{\alpha p / 2} &\left(2 \pi\right)^{\alpha(p-1)/2} \int \exp\left(- (\frac{\alpha}{4\eta} - (\alpha - 1)\frac{M}{2}) \ltwosq{\theta - \theta^*}\right) d\theta\\
= \left(\frac{1}{2\pi}\right)^{\alpha/2} &\left(\frac{1}{2\eta}\right)^{\alpha p / 2} \int \exp\left(- (\frac{\alpha}{4\eta} - (\alpha - 1)\frac{M}{2}) \ltwosq{\theta - \theta^*}\right) d\theta\\
\stackrel{(\ast)}{=} \left(\frac{1}{2\pi}\right)^{\alpha/2} &\left(\frac{1}{2\eta}\right)^{\alpha p / 2} \left(\frac{\frac{\alpha}{4\eta} - (\alpha - 1)\frac{M}{2}}{\pi}\right)^{p/2}\\
&\leq \left(\frac{1}{2\pi}\right)^{\alpha/2} \left(\frac{1}{2\eta}\right)^{\alpha p / 2} \left(\frac{\alpha}{4\pi\eta}\right)^{p/2}\\
\implies D_\alpha(P_3|| P_4) &\leq \frac{\alpha}{\alpha-1}\left(\frac{p}{2} \ln(1/\eta) - \ln(2\pi)\right) + \frac{p}{2} \ln (\alpha/4\pi\eta). 
\end{align}

In $(\ast)$, we use the fact that $\alpha / 4 \eta > (\alpha - 1)M/2$ to ensure the integral converges.

\end{proof}

\begin{lem}\label{lem:pointstart}
Fix some point $\theta_0$. Let $P$ be the distribution $\theta_\eta$ that is the solution to \eqref{eq:ld} from $\theta_0$ for time $\eta \leq 1/2M$. Let $Q$ be the stationary distribution of \eqref{eq:ld}. Then:
\[D_\alpha(\calP || \calQ) = O\Biggl(M^2 \ln(\alpha) \cdot \max\{p \eta^2, \ltwo{\theta_0- \theta^*}^2 \eta^3\}\]
\[ + \frac{\alpha\ltwo{\theta_0-\theta^*}^2}{\eta} + p \ln (\alpha/\eta).\Biggr)\]
\end{lem}
\begin{proof}
By monotonicity of R\'enyi divergences (see e.g., Proposition 9 of \cite{mironov2017renyi}), we can assume $\alpha \geq 2$. Then by applying twice the approximate triangle inequality for R\'enyi divergences (see e.g. Proposition 11 of \cite{mironov2017renyi}), we get:
\[D_\alpha(P_1|| P_4) \leq \frac{5}{3} D_{3\alpha}(P_1|| P_2) + \frac{4}{3} D_{3\alpha-1}(P_2|| P_3) + D_{3\alpha-2}(P_3|| P_4).\]

The lemma now follows by Lemmas~\ref{lem:p1p2divergence},~\ref{lem:p2p3divergence},~\ref{lem:p3p4divergence}.
\end{proof}
Lemma~\ref{lem:renyifrompoint} now follows by plugging $\alpha = 2, \eta = 1/2M$ into Lemma~\ref{lem:pointstart} and then using Theorem 2 of \cite{VempalaWibisono}.

\subsection{Extending to DP-SGLD}
\label{sec:sgldextension}

While we presented our results in terms of DP-LD to simplify the results, a similar result can be proven for DP-SGLD, which is a discrete algorithm and just a reparameterization of DP-SGD, the algorithm we use in our experiments. So, our results can still be applied to some practical settings. We discuss how to modify the proof of Theorem~\ref{thm:checkpointvariance} here.

The only part of the proof of Theorem~\ref{thm:checkpointvariance} which does not immediately hold (or hold in an analogous form) for DP-SGLD is Lemma~\ref{lem:renyifrompoint}. That is, if we can show that starting from a point distribution, we converge to the stationary distribution of DP-LD in a given number of iterations of DP-SGLD, then we can prove an analog of Lemma~\ref{lem:renyifrompoint} and the rest of the proof of Theorem~\ref{thm:checkpointvariance} can be used as-is. 

To prove an analog of Lemma~\ref{lem:renyifrompoint}, we need (i) an analog of Lemma~\ref{lem:pointstart}, which shows that from a point distribution we reach a finite Renyi divergence from the stationary distribution and (ii) an analog of Theorem 2 of \cite{VempalaWibisono}, which shows that from a finite Renyi divergence bound we can reach a small Renyi divergence bound in a given amount of time. 

(i) Can be proven similarly to Lemma~\ref{lem:pointstart}; in particular, we only need Lemmas~\ref{lem:p2p3divergence} and \ref{lem:p3p4divergence}, which by triangle inequality give a Renyi divergence bound between the distribution given after one iteration of DP-SGLD from a point distribution and the stationary distribution. (ii) can be proven using e.g. Lemma 7 of \cite{Erdogdu2020ConvergenceOL}, which shows how the Renyi divergence decreases in every iteration under the assumptions in this section. Getting an exact lower bound on the number of iterations of DP-SGLD needed analogous to our lower bounds on $\kappa_1, \kappa_2$ requires a bit of technical work and results in a much more complicated bound than Theorem~\ref{thm:checkpointvariance}, so we omit the details here. However, we note that an analogous version of one of our high-level takeaways from Theorem~\ref{thm:checkpointvariance}, that $\kappa_1$ can be much larger than $\kappa_2$ in the worst case, would hold for the bounds we could prove for DP-SGLD. In particular, it is still the case that the initial divergence we get from (i) depends on the distance to the minimizer of $\calL$, which can be arbitrarily bad for the initialization but which we can bound with high probability for the intermediate checkpoints via Lemma~\ref{lem:renyifrompoint}.
\section{Missing details from Section~\ref{sec:empEval}}\label{exp_appdx}
Below we provide some preliminaries, details about the experimental setup, and results that were omitted from Section~\ref{sec:empEval} due to space constraints.

\begin{table*}
\caption{Training hyperparameters that we use for StackOverflow experiments with DP-FTRL~\citep{denisov2022improved} and various training (Section~\ref{acc:tr_framework}) and inference (Section~\ref{acc:inf_framework}) aggregations. We use the hyperparameters in "Baseline" section for all the inference aggregations; we discuss how we tune individual parameters of the aggregations in Section~\ref{appdx:hparam_tuning}}
\label{tab:so_hparams}
\centering
\resizebox{\linewidth}{!}{%
\begin{tabular} {|c|c|c|c|c|c|c|c|}
    \hline
    Aggregation & Privacy  & Parameter & clip norm & noise multiplier & server lr & client lr & server momentum \\ \hline 
    \multirow{3}{*}{Baseline} & $\eps = \infty$ & -- & 1.0 & 0.0 & 3.0 & 0.5 & 0.9 \\ \cline{2-8}
    & $\eps = 18.9$ & -- & 1.0 & 0.341 & 0.5 & 1.0 & 0.95 \\ \cline{2-8}
    & $\eps = 8.2$ & -- & 1.0 & 0.682 & 0.25 & 1.0 & 0.95 \\ \hline
    
    \multirow{3}{*}{\algtr{UPA}} & $\eps = \infty$ & $k=3$ & 1.0 & 0.0 & 2.0 & 0.5 & 0.95 \\ \cline{2-8}
    & $\eps = 18.9$ & $k=3$ & 0.3 & 0.341 & 2.0 & 1.0 & 0.95 \\ \cline{2-8}
    & $\eps = 8.2$ & $k=3$ & 0.3 & 0.682 & 1.0 & 1.0 & 0.95 \\ \hline
    
    \multirow{3}{*}{\algtr{EMA}} & $\eps = \infty$ & $\beta=0.95$ & 1.0 & 0.0 & 2.0 & 1.0 & 0.95 \\ \cline{2-8}
    & $\eps = 18.9$ & $\beta=0.95$ & 1.0 & 0.341 & 0.5 & 1.0 & 0.95\\ \cline{2-8}
    & $\eps = 8.2$ & $\beta=0.95$ & 1.0 & 0.682 & 0.25 & 1.0 & 0.95\\
    \hline
\end{tabular}}
\end{table*}

\begin{table*}
\caption{Training hyperparameters that we use for \emph{periodic distribution shifting StackOverflow} experiments with DP-FTRL~\citep{denisov2022improved} and various training (Section~\ref{acc:tr_framework}) and inference (Section~\ref{acc:inf_framework}) aggregations. We use the hyperparameters in "Baseline" section for all the inference aggregations; we discuss how we tune individual parameters of the aggregations in Section~\ref{appdx:hparam_tuning}}
\label{tab:pds_so_hparams}
\centering
\resizebox{\linewidth}{!}{%
\begin{tabular} {|c|c|c|c|c|c|c|c|}
    \hline
    Aggregation & Privacy $\eps$  & Parameter & clip norm & noise multiplier & server lr & client lr & server momentum \\ \hline 
    \multirow{3}{*}{Baseline} & $\infty$ & -- & 1.0 & 0.0 & 3.0 & 0.5 & 0.9 \\ \cline{2-8}
    & $18.9$ & -- & 1.0 & 0.341 & 0.5 & 1.0 & 0.95 \\ \cline{2-8}
    & $8.2$ & -- & 1.0 & 0.682 & 0.25 & 1.0 & 0.95 \\ \hline
    
    \multirow{3}{*}{\algtr{UPA}} & $\infty$ & $k=5$ & 1.0 & 0.0 & 2.0 & 0.5 & 0.95 \\ \cline{2-8}
    & $18.9$ & $k=5$ & 1.0 & 0.341 & 0.5 & 1.0 & 0.95 \\ \cline{2-8}
    & $8.2$ & $k=5$ & 0.3 & 0.682 & 1.0 & 0.5 & 0.95 \\ \hline
    
    \multirow{3}{*}{\algtr{EMA}} & $\infty$ & $\beta=0.95$ & 1.0 & 0.0 & 2.0 & 0.5 & 0.95 \\ \cline{2-8}
    & $18.9$ & $\beta=0.95$ & 1.0 & 0.341 & 0.5 & 1.0 & 0.95\\ \cline{2-8}
    & $8.2$ & $\beta=0.95$ & 1.0 & 0.682 & 1.0 & 0.5 & 0.95\\
    \hline
\end{tabular}}
\end{table*}

\begin{table*}
\caption{Training hyperparameters that we use for CIFAR10 and periodic distribution shifting (PDS) CIFAR10 experiments with DP-SGD~\citep{denisov2022improved} and various training aggregations (Section~\ref{acc:tr_framework}); we discuss how we tune individual parameters of the aggregations in Section~\ref{appdx:hparam_tuning}}
\label{tab:c10_hparams}
\centering
\begin{tabular} {|c|c|c|c|c|c|}
    \hline
    Aggregation & Privacy  & Parameter & noise multiplier & learning rate & $\tau$ ($T$) \\ \hline 
    \multicolumn{6}{|c|}{CIFAR10; DP-SGD; sample-level privacy} \\ \hline
    \multirow{2}{*}{\algtr{UTA}} & $\eps = 8$ & $k=2$ & 3.0 & 4.0 & 2000 (3068) \\ \cline{2-6}
    & $\eps = 1$ & $k=2$ & 8.0 & 2.0 & 400 (568) \\ \hline
    \multirow{2}{*}{\algtr{EMA}} & $\eps = 8$ & $\beta=0.6$ & 4.0 & 2.0 & 2000 (4559) \\ \cline{2-6}
    & $\eps = 1$ & $\beta=0.5$ & 10.0 & 2.0 & 400 (875)\\ \hline
    
    \multicolumn{6}{|c|}{PDS CIFAR10; DP-SGD; sample-level privacy} \\ \hline
    \multirow{2}{*}{\algtr{UTA}} & $\eps = 8$ & $k=5$ & 3.0 & 2.0 & 2000 (2480) \\ \cline{2-6}
    & $\eps = 1$ & $k=3$ & 8.0 & 2.0 & 400 (460) \\ \hline
    \multirow{2}{*}{\algtr{EMA}} & $\eps = 8$ & $\beta=0.6$ & 3.0 & 2.0 & 1500 (2480)\\ \cline{2-6}
    & $\eps = 1$ &$\beta=0.6$ & 8.0 & 2.0 & 200 (460)\\ \hline
\end{tabular}
\end{table*}

\begin{table*}
\caption{Training hyperparameters that we use for CIFAR100 and periodic distribution shifting (PDS) CIFAR100 experiments with DP-SGD~\citep{denisov2022improved} and various training aggregations (Section~\ref{acc:tr_framework}); we discuss how we tune individual parameters of the aggregations in Section~\ref{appdx:hparam_tuning}}
\label{tab:c100_hparams}
\centering
\begin{tabular} {|c|c|c|c|c|c|}
    \hline
    Aggregation & Privacy  & Parameter & noise multiplier & learning rate & $\tau$ ($T$) \\ \hline 
    \multicolumn{6}{|c|}{CIFAR100; DP-SGD; sample-level privacy} \\ \hline
    \multirow{2}{*}{\algtr{UTA}} & $\eps = 8$ & $k=50$ & 9.4 & 4.0 & 400 (2000) \\ \cline{2-6}
    & $\eps = 1$ & $k=50$ & 21.1 & 4.0 & 100 (250) \\ \hline
    \multirow{2}{*}{\algtr{EMA}} & $\eps = 8$ & $\beta=0.85$ & 9.4 & 4.0 & 1500 (2000) \\ \cline{2-6}
    & $\eps = 1$ & $\beta=0.99$ & 21.1 & 4.0 & 200 (250)\\ \hline
    
    \multicolumn{6}{|c|}{PDS CIFAR100; DP-SGD; sample-level privacy} \\ \hline
    \multirow{2}{*}{\algtr{UTA}} & $\eps = 8$ & $k=10$ & 9.4 & 4.0 & 50 (2000) \\ \cline{2-6}
    & $\eps = 1$ & $k=5$ & 21.1 & 4.0 & 200 (250) \\ \hline
    \multirow{2}{*}{\algtr{EMA}} & $\eps = 8$ & $\beta=0.85$ & 9.4 & 4.0 & 200 (2000)\\ \cline{2-6}
    & $\eps = 1$ &$\beta=0.85$ & 21.1 & 4.0 & 200 (250)\\ \hline
\end{tabular}
\end{table*}

\end{document}